\documentclass[lettersize,journal]{IEEEtran}
\usepackage{amsmath,amsfonts}
\usepackage{array}
\usepackage[caption=false,font=normalsize,labelfont=sf,textfont=sf]{subfig}
\usepackage{textcomp}
\usepackage{url}
\usepackage{verbatim}
\usepackage{graphicx}

\usepackage{tabularx}
\usepackage{xcolor,colortbl}
\usepackage{cite}
\usepackage{float}
\usepackage{algpseudocode}
\usepackage{multirow}
\usepackage{algorithm}

\usepackage{hhline} 
\usepackage{pifont}
\usepackage{makecell}
\usepackage{dblfloatfix}
\usepackage{adjustbox}
\usepackage{cleveref}
\usepackage{arydshln}

\definecolor{lb}{rgb}{0.905, 0.941, 0.972}
\definecolor{mb}{rgb}{0.709, 0.778, 0.884}
\definecolor{db}{rgb}{0.513, 0.615, 0.819}

\algdef{SE}[SUBALG]{Indent}{EndIndent}{}{\algorithmicend\ }%
\algtext*{Indent}
\algtext*{EndIndent}

\def\BibTeX{{\rm B\kern-.05em{\sc i\kern-.025em b}\kern-.08em
    T\kern-.1667em\lower.7ex\hbox{E}\kern-.125emX}}
\usepackage{balance}
\begin{document}
\title{DenseSplat: Densifying Gaussian Splatting SLAM with Neural Radiance Prior}

\author{Mingrui Li$^{*}$, Shuhong Liu$^{*}$ \thanks{$^{*}$ First two authors have contributed equally to this work.}, Tianchen Deng, Hongyu Wang$^{\dag}$ \thanks{$^{\dag}$ Corresponding author: whyu@dlut.edu.cn}
\thanks{Mingrui Li and Hongyu Wang are with the Department of Computer Science, Dalian University of Technology, Dalian, China. Shuhong Liu is with the Department of Mechano-informatics, Information Science and Technology, The University of Tokyo, Tokyo, Japan. Tianchen Deng is with Institute of Medical Robotics and Department of Automation, Shanghai Jiao Tong University, and Key Laboratory of System Control and Information Processing, Ministry of Education. This work was partially supported by JST SPRING, Grant Number JPMJSP2108.}}

\markboth{IEEE Transactions on Visualization and Computer Graphics}%
{How to Use the IEEEtran \LaTeX \ Templates}

\maketitle

\begin{abstract}
Gaussian SLAM systems excel in real-time rendering and fine-grained reconstruction compared to NeRF-based systems. However, their reliance on extensive keyframes is impractical for deployment in real-world robotic systems, which typically operate under sparse-view conditions that can result in substantial holes in the map. To address these challenges, we introduce DenseSplat, the first SLAM system that effectively combines the advantages of NeRF and 3DGS. DenseSplat utilizes sparse keyframes and NeRF priors for initializing primitives that densely populate maps and seamlessly fill gaps. It also implements geometry-aware primitive sampling and pruning strategies to manage granularity and enhance rendering efficiency. Moreover, DenseSplat integrates loop closure and bundle adjustment, significantly enhancing frame-to-frame tracking accuracy. Extensive experiments on multiple large-scale datasets demonstrate that DenseSplat achieves superior performance in tracking and mapping compared to current state-of-the-art methods.
\end{abstract}


\section{Introduction}
\label{sec:intro}
Visual Dense Simultaneous Localization and Mapping (SLAM) is a core area of study in 3D computer vision, focusing on real-time localization of the camera and generating a fine-grained map of the surrounding environment. It plays a crucial role in robot localization and navigation, autonomous vehicles, and Virtual/Augmented Reality (VR/AR).

Recent breakthroughs in differential rendering, specifically with Neural Radiance Fields (NeRF) \cite{mildenhall2021nerf,martin2021nerf,liu2025i2nerf} and 3D Gaussian Splatting (3DGS) \cite{kerbl20233d,kulhanek2024wildgaussians,liu2024deraings}, have significantly advanced visual dense SLAM systems \cite{schops2019bad, huang2021di, runz2017co, bloesch2018codeslam, craig2004tandem, teed2021droid}. NeRF-based neural implicit SLAM \cite{yu2021pixelnerf, barron2021mip, deng2022depth, guo2022nerfren, zhai2024nis} integrates the NeRF model for simultaneous tracking and mapping, which facilitates high-quality, dense online map reconstruction and remarkably improves geometric accuracy. Building on this, Gaussian-based SLAM systems \cite{yan2024gs, keetha2024splatam, matsuki2024gaussian, huang2024photo, peng2024rtg, liu2024structure, deng2024compact} have pushed the boundaries by delivering higher-fidelity map reconstruction. By utilizing explicit Gaussian primitives, 3DGS provides substantial benefits in terms of detailed texture representation \cite{keetha2024splatam}, explicit scene manipulation \cite{li2025sgs}, and remarkable real-time rendering capabilities \cite{deng2024compact}.

\begin{figure}[ht]
\centering
\includegraphics[width=0.46\textwidth]{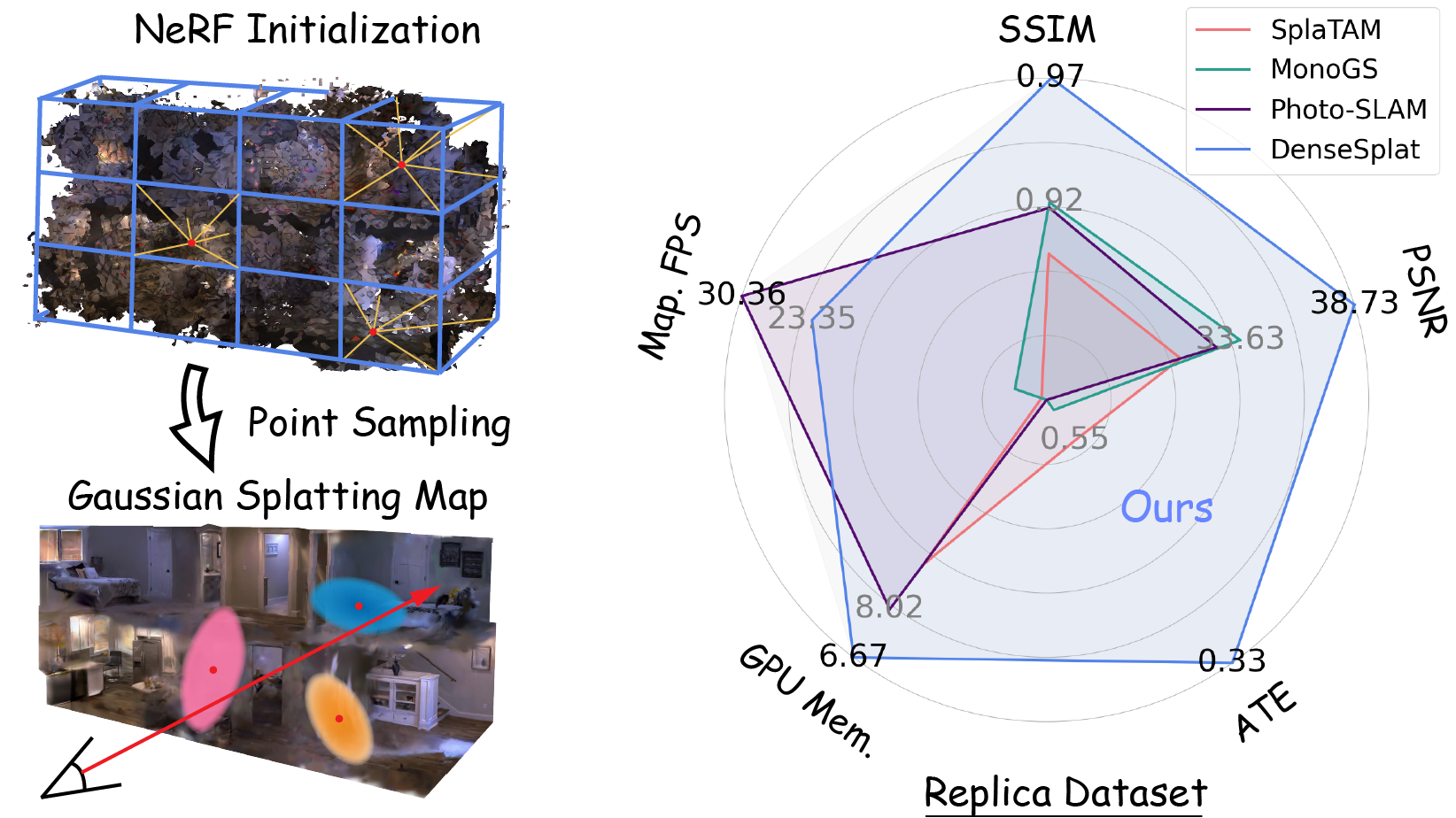}
\caption{DenseSplat leverages NeRF priors into the Gaussian SLAM system, offering superior tracking, fine-grained mapping, and extraordinary real-time performance using sparse keyframes.}
\label{fig:abstract}
\end{figure}

\begin{figure*}[!tp]
\centering
\includegraphics[width=0.9\textwidth]{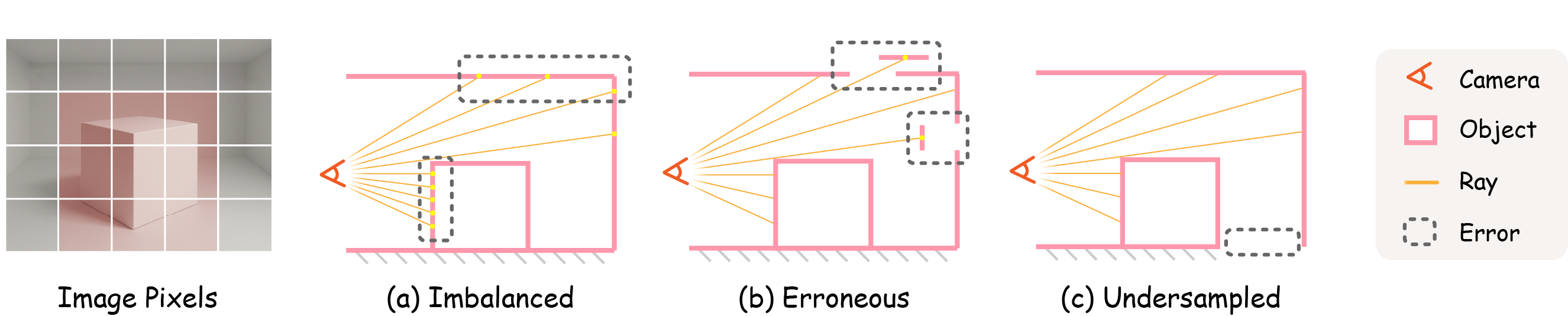}
\caption{Gaussian primitives initialized by direct backprojection from RGB-D streams suffer from several drawbacks. The left pixel image illustrates that closer objects occupy more pixels, whereas farther regions receive fewer rays. (a) Imbalanced point sampling occurs due to the uneven distribution of rays. (b) Erroneous projections arise from inaccurate depth estimates in distant regions. (c) Undersampled areas result from occlusions or insufficient views.}
\label{fig:image-plane}
\end{figure*}

However, 3DGS faces particular challenges within the context of SLAM systems. Unlike offline reconstruction often applied to exhaustive image collections \cite{Knapitsch2017,barron2021mip,liu2025realx3d}, online SLAM systems are typically deployed in complex environments with insufficient observations, leading to partially observed or obstructed views. This shortage significantly affects the completeness of the Gaussian map, as it struggles to effectively interpolate missing geometry of unobserved areas. Moreover, current Gaussian SLAM systems rely on per-frame pixel backprojection from the input stream, which fails to capture the 3D structure of the scene. Such 2D-based initialization results in a map representation where foreground elements are detailed with denser primitives, while the backgrounds, encompassing broader scene geometries, are less defined. \Cref{fig:image-plane} illustrates these limitations, where imbalanced point distribution, erroneous depth projection, and undersampling due to occlusions lead to an uneven and challenging optimization landscape for multiview geometry \cite{niemeyer2024radsplat}. Given these limitations, current Gaussian SLAM systems often maintain a dense keyframe list, such as one in every four frames, during the mapping process that attempts to engage more viewpoints for reliable map optimization \cite{ha2024rgbd}. This reliance on a dense keyframe list demands extensive memory and reduces real-time processing capabilities, both of which are crucial for online SLAM systems.

To address these challenges, we propose employing NeRF priors to densify the Gaussian SLAM system. This densification is achieved through two primary mechanisms using the NeRF model: (1) its interpolation capabilities efficiently close the gaps in the map with densely positioned Gaussian primitives, and (2) it offers a robust initialization of Gaussian primitives that can be densified using extremely sparse keyframes. Moreover, NeRF-based sampling ensures an even distribution of Gaussian primitives aligned with the scene geometry and allows for manageable granularity in the scene representation through specific sampling ratios. Coupled with loop closure and bundle adjustment (BA), our method delivers state-of-the-art performance across large-scale synthetic and real-world datasets as illustrated in \Cref{fig:abstract}. Overall, our contributions can be summarized as follows:

\begin{itemize}
    \item We propose DenseSplat, the first SLAM system combining the advantages of NeRF and 3DGS, capable of real-time tracking, mapping, and loop correction using only sparsely sampled keyframes.
    \item By leveraging NeRF priors, DenseSplat effectively fills the gaps from unobserved or obstructed viewpoints.
    \item We implement geometry-aware primitive sampling and pruning strategies that control the granularity of the Gaussian representation and prune inactive primitives, ensuring a high-fidelity map and real-time processing.
    \item We compare our method with state-of-the-art (SOTA) approaches on multiple datasets, including a challenging large-scale apartment dataset, where DenseSplat achieves superior performance.
\end{itemize}

\section{Related Work}

\subsection{Neural Dense SLAM}
In contrast to traditional SLAM systems \cite{mur2015orb,mur2017orb,zhang2019hierarchical,campos2021orb,du2022accurate,chung2023orbeez,pan2024robust} that utilize point-clouds or voxels for sparse map representation, neural dense SLAM systems \cite{yan2017dense,zhang2023go,li2023end,huang2023real,deng2024plgslam,zhu2024nicer,zhou2024mod,deng2024neslam,xie2025depth,liu2025mg} offer substantial advantages through their dense neural radiance maps, providing a robust foundation for downstream tasks in robotics and AR/XR \cite{deng2023prosgnerf}. iMAP \cite{sucar2021imap} pioneered neural implicit SLAM but suffers from large tracking and mapping error using a single MLP. NICE-SLAM \cite{zhu2022nice} employs multiple MLPs for coarser-to-finer mapping, effectively filling the gaps in reconstruction. ESLAM \cite{johari2023eslam} leverages tri-plane features for efficient scene representation, while Co-SLAM \cite{wang2023co} employs multi-resolution hash-grids for real-time performance. Point-SLAM \cite{sandstrom2023point} relies on neural point clouds for dense scene reconstruction, and Loopy-SLAM \cite{liso2024loopy} introduces map corrections to address scene drift caused by accumulated tracking errors. However, maps reconstructed using NeRF typically lack the quality seen in more recent systems built upon 3D Gaussian Splatting. Furthermore, high-resolution NeRF models often require extensive training and exhibit slower real-time rendering performance \cite{kerbl20233d}, which significantly reduces their practical efficiency.

\subsection{Gaussian Splatting SLAM}

Propelled by advancements in 3DGS \cite{kerbl20233d}, recent Gaussian SLAM systems \cite{yan2024gs,ha2024rgbd,li2025sgs,deng2024compact,hu2025cg} have shown remarkable capabilities in high-fidelity map reconstruction and efficient real-time rendering. Notably, SplaTAM \cite{keetha2024splatam} utilizes isotropic Gaussian representation coupled with dense point-cloud sampling to ensure geometric precision. Conversely, MonoGS \cite{matsuki2024gaussian} employs anisotropic Gaussians to accelerate map reconstruction and enhance texture rendering. Nevertheless, the frame-to-frame tracking mechanisms of these systems do not incorporate loop closure or bundle adjustment, which leads to significant tracking discrepancies in real-world environments. Concurrently, Photo-SLAM \cite{huang2024photo} and RTG-SLAM \cite{peng2024rtg}, which integrate feature-based tracking \cite{mur2017orb,campos2021orb} with dense Gaussian maps, achieve superior tracking accuracy and real-time performance. However, they take the trade-off of diminished rendering quality due to the sparse sampling of scene representation. Moreover, systems such as Gaussian-SLAM \cite{yugay2023gaussian} and LoopSplat \cite{zhu2024loopsplat} propose to implement submap division and fusion strategies to tackle the high memory consumption of fine-grained Gaussian maps. While Gaussian SLAM systems offer superior rendering quality compared to NeRF models, the explicit and discrete nature of their scene representation often results in substantial gaps and holes in the reconstructed map due to unobserved or obstructed views commonly seen in online systems. These deficiencies severely affect their efficiency in real-world applications.

\subsection{Gaussian Splatting with Neural Radiance Prior}

Vanilla 3DGS \cite{kerbl20233d} utilizes the sparse point cloud derived from Structure-from-Motion \cite{bregler2000recovering,schonberger2016structure}, whose reconstruction quality heavily relies on the accuracy of the initial point cloud \cite{jung2024relaxing}. As an alternative, RadSplat \cite{niemeyer2024radsplat} introduces NeRF model \cite{mildenhall2021nerf} into the Gaussian Splatting framework, aiming to achieve robust real-time rendering of complex real-world scenes. RadSplat \cite{niemeyer2024radsplat} specifically employs NeRF as a prior for initialization and supervision, enabling fast training convergence and enhanced quality.

\begin{figure*}[!tp]
\centering
\includegraphics[width=1\textwidth]{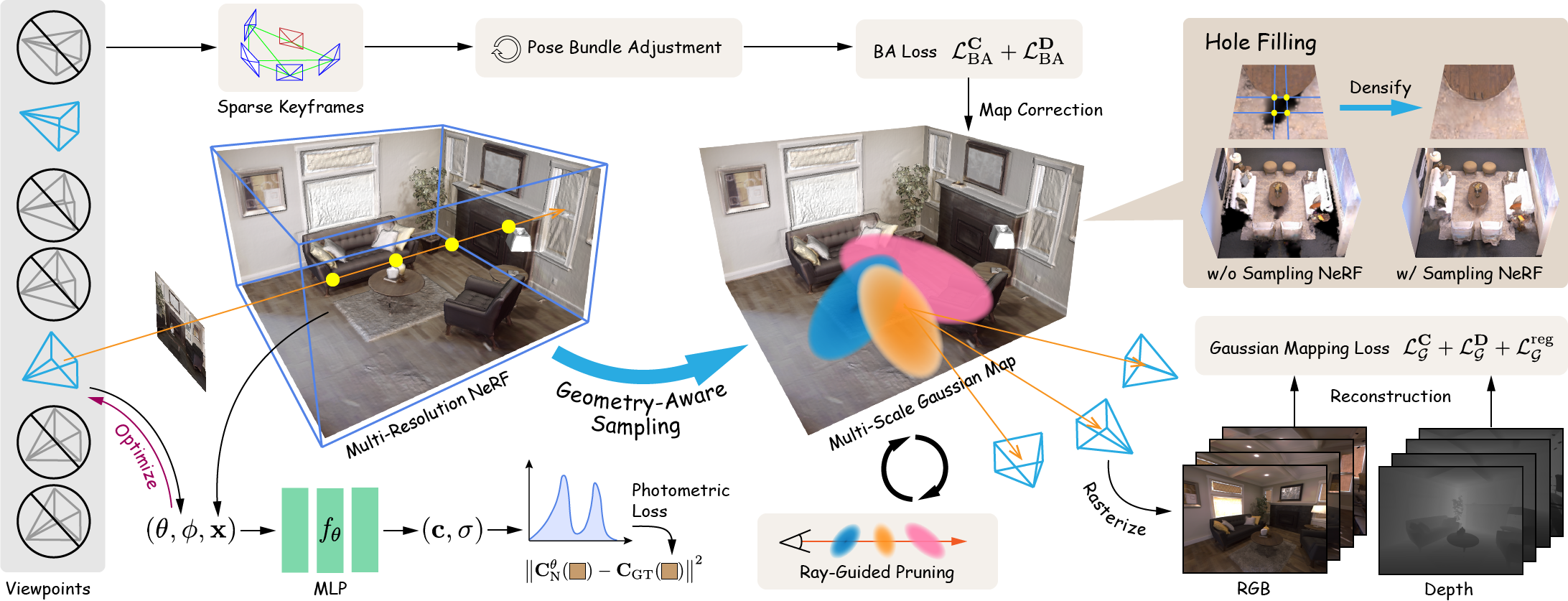}
\caption{{DenseSplat comprises tracking and mapping modules. The tracking module computes camera poses by optimizing the NeRF model and streaming sparse keyframes to the mapping module. Gaussian primitives are produced via geometry-aware sampling, effectively capturing the scene geometry and seamlessly filling gaps. Enhanced by BA-induced map refinement and ray-guided Gaussian pruning strategies, DenseSplat delivers high-quality reconstructions at remarkable real-time speeds.}}
\label{fig:pipline}
\end{figure*}

Compared to RadSplat \cite{niemeyer2024radsplat}, our method differs fundamentally in task objectives and methods. RadSplat \cite{niemeyer2024radsplat} focuses on offline reconstruction scenarios where datasets typically consist of 360-degree or dense viewpoint coverage \cite{barron2022mip,Knapitsch2017}, making completeness and rendering quality the primary goals. In contrast, our method addresses the challenges of real-time SLAM systems deployed in robotic systems, where the limited sensor field of view and navigation path lead to sparse views and sequential observations. These factors frequently result in unobserved or partially observed geometry due to obstacles, posing significant challenges for explicit Gaussian representations that often leave critical gaps in the reconstruction. To overcome these issues, we leverage the NeRF prior not only for initialization but as a robust interpolation mechanism to interpolate unobserved regions and enable real-time adaptability in environments with sparse views. Moreover, unlike RadSplat \cite{niemeyer2024radsplat} that naively samples one million points from all cast rays, we propose a geometry-aware sampling strategy that operates directly in 3D space. This approach uniformly allocates primitives across object surfaces, effectively mitigating the imbalanced sampling caused by foreground bias.

\section{Methods}
\label{headings}

\Cref{fig:pipline} illustrates the overall pipeline of DenseSplat. Starting with an RGB-D stream $\{ I_i, D_i \}_{i=1}^{N}$ of $N$ frames, tracking is initiated by simultaneously optimizing the camera pose and the neural radiance fields $f_{\theta}$ (\Cref{sec:neural-radiance-field}). We then initialize Gaussian primitives guided by points sampled from the implicit radiance fields for fine-grained map reconstruction and scene interpolation (\Cref{sec:gaussian-primitive}). To mitigate drift errors, we implement a local loop closure detection and bundle optimization strategy on the Gaussian map (\Cref{sec:tracking}). Finally, \Cref{sec:map-objective} explains the overall mapping loss and our submap division strategy that effectively reduces memory consumption in the system. Further details on each component are discussed in the subsequent sections.

\subsection{Neural Radiance Prior}
\label{sec:neural-radiance-field}

\noindent \textbf{Preliminaries of Neural Radiance Rendering} The NeRF model $f_\theta$~\cite{mildenhall2021nerf} is a continuous function that predicts colors $\mathbf{c} \in \mathbb{R}^3$ and volume density $\sigma \in \mathbb{R}_+$ along the sampled rays $\mathbf{r}$. Specifically, given a camera origin $\mathbf{o} \in \mathbb{R}^3$ and ray direction $\mathbf{v}$, by uniformly sampling points $\mathbf{x} = \mathbf{o} + \mathbf{d}_{j} \mathbf{v} |_{j \in \{1...\}}$, the pixel color $\mathbf{C}_{\mathrm{N}}$ can be rendered by NeRF using ray marching:
\begin{equation} \label{eq:nerf_render}
    \mathbf{C}_{\mathrm{N}} = \sum_{j=1} \mathbf{c}_j \alpha_j T_{Nj}, \quad \text{and} \quad T_{Nj} = \prod_{t=1}^{j-1}(1-\alpha_t)~,
\end{equation}
where $T_{Ni}$ denotes the transmittance and $\alpha_i = 1-e^{-\sigma_i \delta_i}$ is the alpha values at point $\mathbf{x}_i$. $\delta_j$ represents the spacing between successive sample points. The radiance filed $f_\theta$, parameterized as an MLP with ReLU activation, is trained using gradient descent to minimize the photometric loss:
\begin{equation} \label{eq:nerf-objective}
    \mathcal{L}_{\mathrm{N}} = \sum_{\mathbf{r} \in \mathcal{R}_{\text{b}}}\left\|\mathbf{C}_{\mathrm{N}}^\theta(\mathbf{r}) - \mathbf{C}_{\mathrm{GT}}(\mathbf{r})\right\|^2~.
\end{equation}

\noindent In this context, $\mathbf{r} \in \mathcal{R}_{\text{b}}$ denotes a batch of rays drawn from the entire set of rays that have valid depth measurements, $\mathbf{C}_{\mathrm{GT}}$ is the ground-truth color. \\

\noindent \textbf{NeRF-based Camera Tracking} We track the camera-to-world transformation matrix $\mathbf{T}_{wc} \in \mathrm{SE}(3)$ by optimizing the pose using the objective function of NeRF defined in \Cref{eq:nerf-objective}. The pose is initialized based on the principle of constant velocity:
\begin{equation}
   \mathbf{T}_i = \mathbf{T}_{i-1} \mathbf{T}^{-1}_{i-2} \mathbf{T}_{i-1}~.
\end{equation}

\noindent To enable tracking, we first bootstrap the NeRF model using a small number of initial frames, whose poses are estimated via the constant velocity assumption. These early frames provide rough supervision to begin learning coarse scene geometry and appearance. Once initialized, the NeRF is incrementally updated as more frames are integrated. For each new frame $i$, a two-stage optimization strategy is adopted following the conventional NeRF-based SLAM system \cite{wang2023co}. We begin with pose-only optimization while keeping the NeRF parameters fixed. After the pose stabilizes, we perform joint optimization of both the camera pose and NeRF weights. The NeRF model is used exclusively for per-frame camera tracking. Once tracking is complete, loop closure and bundle adjustment are applied based on the fine-grained Gaussian map to correct cumulative drift, as detailed in \Cref{sec:tracking}. \\

\noindent \textbf{Geometry-aware Point Sampling} Compared to explicit Gaussian representations that require dense viewpoints for thorough scene optimization, NeRF-based models offer remarkable advantages in terms of interpolation, as they can deduce unseen geometry~\cite{pumarola2021d,zhu2022nice,yang2023nerfvs}. To harness this capability for efficient real-time performance, we utilize a multi-resolution hash radiance field as proposed by \cite{muller2022instant}. This grid system enables precise interrogation of volumetric data across various resolutions, effectively capturing detailed surface geometries even in sparsely sampled areas. We identify critical surface transitions by setting a density threshold, $\tau_{\rm grid}$, which examines each grid edge to detect transitions where the density at one corner surpasses $\tau_{\rm grid}$, while at its adjacent corner, it does not. At these transitions, we interpolate the surface crossing points using:
\begin{equation}\label{eq:grid-sample}
    \mathbf{x} = \mathbf{x}_1 + \frac{\tau_{\rm grid} - \sigma(\mathbf{x}_1)}{\sigma(\mathbf{x}_2) - \sigma(\mathbf{x}_1)} \cdot (\mathbf{x}_2 - \mathbf{x}_1)~.
\end{equation}

\noindent Here, $\mathbf{x}_1$ and $\mathbf{x}_2$ represent sampled grid points, and $\sigma$ denotes volumn density. We then compile these points into a point cloud. This approach leverages the robust interpolation capabilities of NeRF to provide dense and geometry-aware initialization of Gaussian primitives, which are further refined in subsequent mapping steps.

\subsection{Fine-grained Gaussian Map}
\label{sec:gaussian-primitive}

\noindent \textbf{Multi-scale Gaussian Rendering} By initializing through grid sampling from th NeRF model, we represent the scene using a set of anisotropic 3D Gaussian primitives ${\{\mathcal{G}_j\}}$~\cite{kerbl20233d}. Each primitive $\mathcal{G}_j$ is defined by a mean $\mu_j \in \mathbb{R}^3$, a covariance matrix $\Sigma_j$, an opacity value $\alpha_j \in [0,1]$, coefficients for third-order spherical harmonics $\mathbf{SH}_j \in \mathbb{R}^{16}$, and a scaling factor $s_j \in \mathbb{R}^3$. During the rendering process, these primitives are first projected onto a 2D plane, transforming them into 2D Gaussians. The transformation utilizes a viewing matrix $W$, and the resulting 2D covariance matrix $\Sigma'_j$ in the image space is computed as follows \cite{zwicker2001surface}: \begin{equation}
    \Sigma'_j = (JW\Sigma_jW^TJ^T)_{1:2,1:2}~,
\end{equation}
where $J$ represents the Jacobian of an affine projection approximation. The mean $\mu'_j$ of the 2D Gaussian is derived by projecting $\mu_j$ onto the image plane using $W$. Subsequently, these projected Gaussians are sorted from the nearest to the farthest and rendered using an alpha-blending process akin to \Cref{eq:nerf_render}. This results in rasterized pixel color $\mathbf{C}_{\mathcal{G}}$ and depth value $\mathbf{D}_{\mathcal{G}}$:
\begin{equation}
    \mathbf{C}_{\mathcal{G}} = \sum_{j=1} \mathbf{c}_j \alpha_j T_{\mathcal{G}j} \quad \text{and} \quad \mathbf{D}_{\mathcal{G}} = \sum_{j=1} \mathbf{d}_j \alpha_j T_{\mathcal{G}j}~,
\end{equation}
where $\alpha_j$ denotes the blending weight, $\mathbf{d}_j$ represents the depth of each Gaussian relative to the image plane, and $T_{\mathcal{G}j}$ is the transmittance, calculated similar to \Cref{eq:nerf_render}, using the opacity $\alpha_j$ of each Gaussian $\mathcal{G}_j$.

The radiance field enables interpolated point initialization for Gaussian primitives, yet it can also lead to aliasing effects that diminish the quality of the map during sampling~\cite{barron2021mip,barron2023zip}. This issue becomes particularly acute at the edges of objects where small Gaussian floaters result in significant artifacts. Drawing inspiration from~\cite{yu2024mip}, we adopt a multi-scale Gaussian rendering strategy that consolidates smaller Gaussians into larger ones to enhance scene consistency. Specifically, we use Gaussian functions across four levels of detail, corresponding to down-sampling resolutions of 1×, 4×, 16×, and 64×. Throughout the training phase, we merge smaller, fine-level Gaussians into larger, coarse-level Gaussians. The selection of Gaussians for merging is determined by pixel coverage, which picks Gaussians based on the coverage range set by the inverse of the highest frequency component in the region, denoted as $f_{\rm max} = 1/s_j$, where $s_j$ is the scaling factor.
\\
 
\noindent \textbf{Ray-Guided Gaussian Pruning} Sampling from the NeRF model may also introduce erroneous Gaussian floaters and outliers that affect the quality of the reconstruction. To reduce redundant Gaussians produced by the densification process and enhance rendering efficiency, we implement a pruning strategy based on sampled rays from the NeRF model. Specifically, we use an importance assessment to identify and remove inactive Gaussians from the map during optimization. The importance of each Gaussian is quantified based on its contribution to sampled rays across all input images $\{I_i\}_{i=1}^{N}$. Drawing inspiration from~\cite{niemeyer2024radsplat}, we implement the score function for each primitive as:
\begin{equation}
\mathcal{E} (\mathcal{G}_i) = \max_{r \in I_i} (\alpha_i^r T_{{\mathcal{G}_i}}^r)~,
\end{equation}
\noindent where $\alpha_i^r T_{\mathcal{G}_i}^r$ captures the Gaussian $\mathcal{G}_i$'s contribution to the final color prediction of a pixel along the ray $r$. We then compute a pruning mask:
\begin{equation}\label{eq:gs-prune}
    M(\mathcal{G}_i) = \mathbf{1}(\mathcal{E}(\mathcal{G}_i) < \tau_{\text{prune}})~,
\end{equation}
\noindent where primitives under a pruning threshold $\tau_{\rm prune} \in [0, 1]$ are removed from the map. Note that the Gaussians initialized by the NeRF model are exempt from this pruning process to avoid removing those that bridge the gaps for obstructed viewpoints and to ensure a manageable granularity.

\subsection{Loop Closure and Bundle Adjustment}
\label{sec:tracking}

In the bundle adjustment (BA) process, we use the Bag of Words (BoW) model~\cite{campos2021orb} to determine the relevance between keyframes. Upon detecting a loop, this triggers a BA procedure for the involved keyframes, in a manner akin to~\cite{liso2024loopy, zhu2024loopsplat}. The implementation is detailed in \Cref{algo:ba}.

\begin{algorithm}[!t]
\caption{Loop Closure and Bundle Adjustment}
\label{algo:ba}
\begin{algorithmic}[1]
\State \textbf{Input:} Keyframe stream \( \{I_k, D_k, \mathbf{T}_k\} \), BoW database \( \mathcal{B} \)
\State \textbf{Output:} Updated keyframe poses \( \{\mathbf{T}_k\} \)

\For{each new keyframe \( (I_k, D_k, \mathbf{T}_k) \)}
    \State \textbf{// Add keyframe to BoW database}
    \State \( \mathcal{B} \leftarrow \mathcal{B} \cup \{I_k\} \)

    \State \textbf{// Retrieve loop closure candidates}
    \State \( \{j\} \leftarrow \text{QueryBoW}(\mathcal{B}, I_k) \)

    \For{each candidate \( j \)}
        \If{LoopGeometricCheck(\( I_k, I_j \))}
            \State \textbf{// Loop closure confirmed}
            \State \( \mathbf{T}_{k \rightarrow j} \leftarrow \text{EstimateRelativePose}(I_k, I_j) \)

            \State \textbf{// Collect covisible keyframes}
            \State \( \mathcal{K}_{\text{local}} \leftarrow \text{GetCovisibleKeyframes}(k, j) \)

            \State \textbf{// Pose-only Bundle Adjustment}
            \State \( \{\mathbf{T}_p\} \leftarrow \text{RunPoseBA}(\mathcal{K}_{\text{local}}) \)

            \State \textbf{// Update keyframe poses}
            \For{each \( p \in \mathcal{K}_{\text{local}} \)}
                \State \( \mathbf{T}_p \leftarrow \text{OptimizedPose}(p) \)
            \EndFor
        \EndIf
    \EndFor
\EndFor
\end{algorithmic}
\end{algorithm}

While BA refines the camera trajectories, the dense map cannot trivially incorporate these updates and remains synchronized with the pre-optimized poses, which leads to spatial drift. To maintain the geometric and visual consistency of the map after the BA process, we adjust the color $\mathbf{C}_{\mathcal{G}} \langle\mathbf{T}_i\rangle$ and depth $\mathbf{D}_{\mathcal{G}}\langle\mathbf{T}_i\rangle$ rendered at current pose $\mathbf{T}_i$ to warp the co-visible keyframe pose $\mathbf{T}_k$ using the estimated relative pose transformation $\langle \mathbf{T}_{i \rightarrow k} \rangle$. We then construct the BA-induced mapping loss using the following equations:
\begin{align}
    \label{eq:ba-rgb}
    \mathcal{L}_{\rm BA}^{\mathbf{C}} &= \sum_{k=1}^{N-1} \sum_{i=k+1}^N \| \mathbf{C}_{\mathcal{G}} \langle \mathbf{T}_{i \rightarrow k} \rangle - \mathbf{C}_{\mathcal{G}} \langle \mathbf{T}_k \rangle \| \\
    \label{eq:ba-depth}
    \mathcal{L}_{\rm BA}^{\mathbf{D}} &= \sum_{k=1}^{N-1} \sum_{i=k+1}^N \| \mathbf{D}_{\mathcal{G}} \langle \mathbf{T}_{i \rightarrow k} \rangle - \mathbf{D}_{\mathcal{G}} \langle \mathbf{T}_k \rangle \|
\end{align}

\noindent Leveraging the rapid rendering capabilities of 3D Gaussian Splatting, our method supports real-time re-rendering and the correction of drift errors.

\subsection{Gaussian Map Optimization}
\label{sec:map-objective}

\noindent \textbf{Mapping Objective Function} In our experiments, we observe that the aggregated Gaussians can experience scale explosion during the BA process, potentially introducing artifacts into the map. To mitigate this issue, we introduce an L2 regularization loss $\mathcal{L}_{\rm reg}$ for Gaussian primitives whose scales exceed a threshold of $\tau_{\rm scale} = 1$. The overall mapping loss is thereby defined as:
\begin{align}
    \mathcal{L}_{\mathcal{G}} &= \lambda_{\rm c} \| \mathbf{C}_{\mathcal{G}} - \mathbf{C}_{\rm GT} \| \nonumber + \lambda_{\rm d} \| \mathbf{D}_{\mathcal{G}} - \mathbf{D}_{\rm GT} \| \\
    &+ \lambda_{\rm ssim} {\rm SSIM} (\mathbf{C}_{\mathcal{G}}, \mathbf{C}_{\rm GT} ) + \lambda_{\rm reg} \mathcal{L}_{\rm reg}~,
\end{align}

\noindent where $\mathbf{C}_{\rm GT}$ and $\mathbf{D}_{\rm GT}$ denote the ground-truth color and depth from the input stream. The SSIM loss~\cite{wang2004image} calculates the structural similarity between the rendered and ground-truth images. The coefficients $\lambda_{\rm c}$, $\lambda_{\rm d}$, $\lambda_{\rm ssim}$, and $\lambda_{\rm reg}$ are weighting hyperparameters. \\

\begin{table*}[tb]
    \centering
    \setlength{\tabcolsep}{0pt}
    \caption{Quantitative comparison of our method and the baselines in training view rendering on the Replica dataset \cite{straub2019replica}. The \underline{Underline} indicates that relocalization was triggered due to accumulated tracking errors in the ORB tracking system~\cite{campos2021orb}. Dash presents system failure.}
    \footnotesize 
    \begin{tabularx}{\textwidth}{@{}@{\hspace{1pt}}c@{\hspace{3pt}}|>{\hspace{2pt}}l@{\hspace{1pt}} | >{\hspace{2pt}}c@{\hspace{2pt}} *{14}{>{\centering\arraybackslash}X}}
        \Xhline{2\arrayrulewidth}
        & \textbf{Methods} & \textbf{Metrics} & room0 & room1 & room2 & office0 & office1 & office2 & office3 & office4 & apart0 & apart1 & apart2 & frl0 & frl4 & \textbf{Avg.} \\
        \hline
        && PSNR$\uparrow$ & 28.88 & 28.51 & 29.37 & 35.44 & 34.63 & 26.56 & 28.79 & 32.16 & 30.10 & 22.86 & 23.29 & 23.52 & 25.33 & 28.42 \\
        \multirow{0}{*}{\rotatebox[origin=c]{90}{NeRF-SLAM}} & Co-SLAM~\cite{wang2023co}~ & SSIM$\uparrow$ & 0.892 & 0.843 & 0.851 & 0.854 & 0.826 & 0.814 & 0.866 & 0.856 & 0.905 & 0.766 & 0.771 & 0.822 & 0.814 & 0.837 \\
        && LPIPS$\downarrow$ & 0.213 & 0.205 & 0.215 & 0.177 & 0.181 & 0.172 & 0.163 & 0.176 & 0.321 & 0.440 & 0.462 & 0.367 & 0.461 & 0.273 \\
        \cline{2-17}
        && PSNR$\uparrow$ & 31.80 & 32.70 & 32.70 & 38.66 & 15.96 & 15.00 & 33.61 & 34.15 & \cellcolor{lb}\textbf{33.71} & 23.74 & 25.96 & 25.55 & \cellcolor{lb}\textbf{34.55} & 29.08 \\
        & Loopy-SLAM~\cite{liso2024loopy}~ & SSIM$\uparrow$ & 0.912 & 0.914 & 0.917 & 0.960 & 0.088 & 0.583 & 0.921 & 0.935 & \cellcolor{lb}\textbf{0.927} & 0.833 & 0.842 & 0.889 & 0.930 & 0.819 \\
        && LPIPS$\downarrow$ & 0.167 & 0.198 & 0.205 & 0.126 & 0.369 & 0.538 & 0.202 & 0.187 & 0.248 & 0.293 & 0.288 & 0.195 & 0.226 & 0.249 \\
        \cline{2-17}
        && PSNR$\uparrow$ & 32.40 & 34.08 & 35.50 & 38.26 & 39.16 & \cellcolor{lb}\textbf{33.99} & \cellcolor{mb}\textbf{33.48} & 33.49 & \cellcolor{mb}\textbf{34.95} & \cellcolor{mb}\textbf{32.27} & \cellcolor{mb}\textbf{33.31} & \cellcolor{mb}\textbf{36.01} & \cellcolor{mb}\textbf{34.87} & \cellcolor{mb}\textbf{34.75} \\
        & Point-SLAM~\cite{sandstrom2023point}~ & SSIM$\uparrow$ & \cellcolor{mb}\textbf{0.974} & \cellcolor{mb}\textbf{0.977} & \cellcolor{db}\textbf{0.982} & 0.983 & \cellcolor{lb}\textbf{0.986} & 0.960 & 0.960 & 0.979 & \cellcolor{mb}\textbf{0.970} & \cellcolor{mb}\textbf{0.929} & \cellcolor{mb}\textbf{0.944} & \cellcolor{mb}\textbf{0.960} & \cellcolor{mb}\textbf{0.970} & \cellcolor{mb}\textbf{0.965} \\
        && LPIPS$\downarrow$ & 0.113 & 0.116 & 0.111 & 0.100 & 0.118 & 0.156 & 0.132 & 0.142 & \cellcolor{mb}\textbf{0.153} & \cellcolor{mb}\textbf{0.205} & \cellcolor{mb}\textbf{0.211} & \cellcolor{lb}\textbf{0.156} & \cellcolor{lb}\textbf{0.176} & 0.147 \\
        \hline
        && PSNR$\uparrow$ & \cellcolor{lb}\textbf{32.49} & 33.72 & 34.65 & 38.29 & 39.04 & 31.91 & 30.05 & 31.83 & 13.12 & 24.57 & 28.52 & 31.82 & 32.71 & 30.98 \\
        & SplaTAM~\cite{keetha2024splatam}~ & SSIM$\uparrow$ & \cellcolor{db}\textbf{0.975} & \cellcolor{lb}\textbf{0.970} & \cellcolor{lb}\textbf{0.980} & 0.982 & 0.982 & \cellcolor{lb}\textbf{0.965} & 0.952 & 0.949 & 0.415 & 0.821 & 0.883 & 0.930 & \cellcolor{lb}\textbf{0.945} & 0.904 \\
        && LPIPS$\downarrow$ & 0.072 & \cellcolor{lb}\textbf{0.096} & 0.078 & 0.086 & 0.093 & 0.100 & 0.110 & 0.150 & 0.656 & 0.302 & 0.241 & 0.164 & 0.184 & 0.179 \\
        \cline{2-17}
        && PSNR$\uparrow$ & 29.57 & 31.61 & 33.46 & 38.39 & 39.62 & 32.91 & 33.62 &  34.26 & 27.07 & 24.93 & 24.34 & 25.19 & 26.70 & 30.90 \\
        & Gauss-SLAM~\cite{yugay2023gaussian}~ & SSIM$\uparrow$ & 0.944 & 0.952 & 0.973 & 0.985 & \cellcolor{mb}\textbf{0.991} & 0.974 & \cellcolor{db}\textbf{0.982} & \cellcolor{lb}\textbf{0.979} & 0.864 & 0.850 & 0.828 & 0.836 & 0.839 & 0.923 \\
        && LPIPS$\downarrow$ & 0.197 & 0.184 & 0.148 & 0.099 & 0.097 & 0.158 & 0.123 & 0.138 & 0.345 & 0.381 & 0.410 & 0.358 & 0.343 & 0.229 \\
        \cline{2-17}
        && PSNR$\uparrow$ & 30.71 & 33.51 & 35.02 & 38.47 & 39.08 & 33.03 &  \cellcolor{lb}\textbf{33.78} & 36.02 & \underline{29.07} & \underline{22.73} & 24.59 & \cellcolor{lb}\textbf{34.16} & 33.36 & 32.58 \\
        \multirow{-1.8}{*}{\rotatebox[origin=c]{90}{Gaussian-SLAM}} & Photo-SLAM~\cite{huang2024photo}~ & SSIM$\uparrow$ & 0.899 & 0.934 &  0.951 & 0.964 & 0.961 & 0.938 & 0.938 & 0.952 & \underline{0.922} & \underline{0.796} & 0.848 & \cellcolor{lb}\textbf{0.940} & 0.932 & 0.921 \\
        && LPIPS$\downarrow$ & \cellcolor{lb}\textbf{0.075} & 0.057 & \cellcolor{db}\textbf{0.043} & \cellcolor{db}\textbf{0.050} & \cellcolor{db}\textbf{0.047} & \cellcolor{mb}\textbf{0.077} & \cellcolor{lb}\textbf{0.066} & \cellcolor{mb}\textbf{0.054} & \cellcolor{lb}\textbf{\underline{0.227}} & \underline{0.293} & 0.354 & \cellcolor{mb}\textbf{0.115} & \cellcolor{mb}\textbf{0.129} & \cellcolor{mb}\textbf{0.122} \\
        \cline{2-17}
        && PSNR$\uparrow$ & \cellcolor{mb}\textbf{34.83} & \cellcolor{mb}\textbf{36.43} & \cellcolor{db}\textbf{37.49} & \cellcolor{mb}\textbf{39.95} & \cellcolor{db}\textbf{42.09} & \cellcolor{mb}\textbf{36.24} & \cellcolor{mb}\textbf{36.70} & \cellcolor{lb}\textbf{36.07} & 22.91 & 26.88 & 27.93 & 31.72 & 27.98 & 33.63 \\
        & MonoGS~\cite{matsuki2024gaussian}~ & SSIM$\uparrow$ & 0.954 & 0.959 & 0.965 & 0.971 & 0.977 & 0.964 & 0.963 & 0.957 & 0.864 & 0.835 & 0.836 & 0.886 & 0.873 & 0.923 \\
        && LPIPS$\downarrow$ & \cellcolor{mb}\textbf{0.068} & \cellcolor{mb}\textbf{0.076} & \cellcolor{lb}\textbf{0.075} & 0.072 & \cellcolor{lb}\textbf{0.055} & \cellcolor{lb}\textbf{0.078} & \cellcolor{mb}\textbf{0.065} & \cellcolor{lb}\textbf{0.099} & 0.385 & 0.284 & 0.272 & 0.225 & 0.245 & 0.154 \\
        \cline{2-17}
        && PSNR$\uparrow$ & 31.56 & \cellcolor{lb}\textbf{34.21} & \cellcolor{lb}\textbf{35.57} & \cellcolor{lb}\textbf{39.11} & \cellcolor{lb}\textbf{40.27} & 33.54 & 32.76 & \cellcolor{mb}\textbf{36.48} & - & \cellcolor{lb}\textbf{29.08} & \cellcolor{lb}\textbf{29.14} & 33.88 & - & \cellcolor{lb}\textbf{34.14} \\
        & RTG-SLAM~\cite{peng2024rtg}~ & SSIM$\uparrow$ & \cellcolor{lb}\textbf{0.967} & \cellcolor{db}\textbf{0.979} & \cellcolor{mb}\textbf{0.981} & \cellcolor{mb}\textbf{0.990} & \cellcolor{db}\textbf{0.992} & \cellcolor{db}\textbf{0.981} & \cellcolor{mb}\textbf{0.981} & \cellcolor{mb}\textbf{0.984} & - & \cellcolor{lb}\textbf{0.900} & \cellcolor{lb}\textbf{0.909} & 0.933 & - & \cellcolor{lb}\textbf{0.963} \\
        && LPIPS$\downarrow$ & 0.131 & 0.105 & 0.115 & \cellcolor{lb}\textbf{0.068} & 0.075 & 0.134 & 0.128 & 0.117 & - & \cellcolor{lb}\textbf{0.232} & \cellcolor{lb}\textbf{0.233} & 0.181 & - & \cellcolor{lb}\textbf{0.138} \\
        \cline{2-17}
        \noalign{\vskip 0.2pt}
        && PSNR$\uparrow$ & \cellcolor{db}\textbf{37.11} & \cellcolor{db}\textbf{36.64} &
        \cellcolor{mb}\textbf{36.32} &
        \cellcolor{db}\textbf{39.97} &
        \cellcolor{mb}\textbf{41.69} &
        \cellcolor{db}\textbf{37.84} &
        \cellcolor{db}\textbf{37.52} &
        \cellcolor{db}\textbf{40.76} &
        \cellcolor{db}\textbf{38.44} &
        \cellcolor{db}\textbf{37.31} &
        \cellcolor{db}\textbf{37.37} &
        \cellcolor{db}\textbf{40.66} &
        \cellcolor{db}\textbf{41.89} &
        \cellcolor{db}\textbf{38.73} \\
        & \textbf{Ours} & SSIM$\uparrow$ &
        0.958 &
        0.954 &
        0.956 &
        \cellcolor{lb}\textbf{0.986} &
        0.977 &
        \cellcolor{mb}\textbf{0.976} &
        \cellcolor{lb}\textbf{0.966} &
        \cellcolor{db}\textbf{0.985} &
        \cellcolor{db}\textbf{0.972} &
        \cellcolor{db}\textbf{0.945} &
        \cellcolor{db}\textbf{0.951} &
        \cellcolor{db}\textbf{0.986} &
        \cellcolor{db}\textbf{0.988} &
        \cellcolor{db}\textbf{0.969} \\
        && LPIPS$\downarrow$ &
        \cellcolor{db}\textbf{0.065} &
        \cellcolor{db}\textbf{0.060} &
        \cellcolor{mb}\textbf{0.062} &
        \cellcolor{mb}\textbf{0.055} &
        \cellcolor{mb}\textbf{0.053} &
        \cellcolor{db}\textbf{0.074} &
        \cellcolor{db}\textbf{0.064} &
        \cellcolor{db}\textbf{0.049} &
        \cellcolor{db}\textbf{0.057} &
        \cellcolor{db}\textbf{0.055} &
        \cellcolor{db}\textbf{0.054} &
        \cellcolor{db}\textbf{0.046} &
        \cellcolor{db}\textbf{0.038} &
        \cellcolor{db}\textbf{0.056} \\
        \Xhline{2\arrayrulewidth}
     \end{tabularx}
    \label{tab:replica_psnr}
\end{table*}

\noindent \textbf{Submap Division and Fusion} When deploying a SLAM system in large-scale environments, managing the excessive memory consumption of the dense mapping is critical for practical applications. To address this issue, we employ an effective submap division and fusion strategy presented in \Cref{algo:submap}. Specifically, we partition input frames into submaps at intervals of every 400 frames, structured as follows:
\begin{equation}
\left\{I_i, D_i\right\}_{i=1}^N \mapsto\left\{\mathrm{SF}_{\langle f_\theta^1, \mathcal{G}^1 \rangle }^1, \mathrm{SF}_{\langle f_\theta^2, \mathcal{G}^2 \rangle}^2, \ldots, \mathrm{SF}_{\langle f_\theta^n, \mathcal{G}^n \rangle}^n\right\}~,
\end{equation}
\noindent where $\mathrm{SF}_{\langle f_\theta, \mathcal{G}\rangle}^n$ represents each submap used to develop NeRF models and subsequent Gaussian maps. While the explicit Gaussian representation enables the seamless combination of submaps into a global map, directly fusing submaps remains a challenging task. Drawing inspiration from Mipsfusion \cite{tang2023mips}, we utilize anchor-frame BA during our submap fusion process to achieve precise alignment and seamless fusion at submap boundaries. Each submap is anchored based on the estimated pose of its first frame. Following BA, we precisely adjust the central pose of each submap to ensure accurate re-anchoring. Our submap strategy effectively reduces memory consumption by enabling the parallel construction of each submap, thereby mitigating the issues associated with the continuous expansion of the global map.


\begin{algorithm}[!t] \caption{Submap Division and Fusion Strategy with Submap Bundle Adjustment}
\label{algo:submap}
\begin{algorithmic}[1]
\State \textbf{Input:} RGB-D sequences \( \{I_i, D_i\} \) and poses \( \{\mathbf{T}_i\} \)
\State \textbf{Output:}
\Indent
\State Submaps \( \{\mathrm{SF}_{\langle f_\theta, \mathcal{G}\rangle}\} \)
\State keyframe set \( \{I_i, D_i, \mathbf{T}_i \} \in \Omega_{\mathrm{SF}}\)
\State anchor-frame set \( \{I_i, D_i, \mathbf{T}_i \} \in \Lambda\)
\EndIndent
\State Initiate frame index \( i \leftarrow 0 \) and submap index \( j \leftarrow 0 \)
\Repeat

\State \textbf{// Insert Keyframes:}
\If{CheckKeyframe(\( \{I_i, D_i, \mathbf{T}_i \} \))}
\State \( \Omega_{\mathrm{SF}^j} \leftarrow \text{InsertKeyFrame}(I_i, D_i, \mathbf{T}_i) \)
\EndIf

\State \textbf{// NeRF Submap Creation:}
\If{CheckSubmapCreation(i)}
\State \( \mathrm{SF}^j_{f_\theta} \leftarrow \text{CreateNeRFSubmap}(I_i, D_i, \mathbf{T}_i) \)
\State \( \Lambda \leftarrow \text{InsertAnchorFrame}(I_i, D_i, \mathbf{T}_i) \)
\EndIf
\State \textbf{// Gaussian Submap Creation:}
\If{Length(\(\Lambda\)) \(>j\)} \State \( \mathrm{SF}^j_{\mathcal{G}} \leftarrow \text{RenderGaussianSubmap}(\mathrm{SF}^j_{f_\theta}, \Omega_{\mathrm{SF}^j}) \)
\State \textbf{// Gaussian Submaps BA:}
\State \( \{\mathrm{SF}_{\langle f_\theta, \mathcal{G}\rangle}\} \leftarrow \text{AnchorFrameBA}(\Lambda, \{\mathrm{SF}^n_{\langle f_\theta, \mathcal{G}\rangle}|_{n\leq j}\})\)
\State \( j \leftarrow j + 1 \)
\EndIf
\State \( i \leftarrow i + 1 \) \Until{All frames are processed}
\end{algorithmic} \end{algorithm}

\section{Experiments}

\begin{figure*}[t]
\centering
\includegraphics[width=\textwidth]{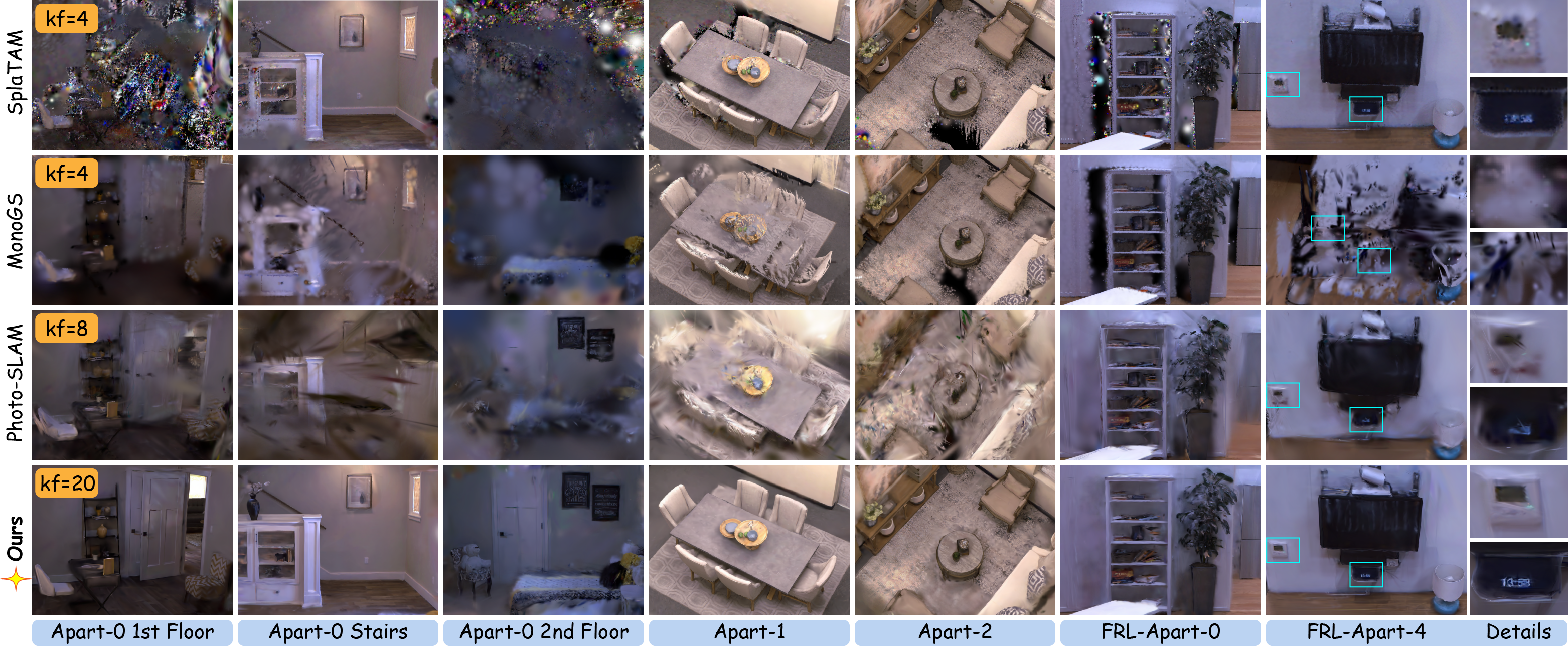}
\caption{\textbf{Novel-view synthesis (NVS)} comparison of DenseSplat and Gaussian-based baseline methods on the selected scenes of Replica apartment dataset~\cite{straub2019replica}. Our method demonstrates superior performance in geometric accuracy, hole filling, and fine-grained texture rendering. Crucially, DenseSplat utilizes a sparse keyframe (kf) interval of 20, offering a more efficient and practical setup compared to the dense keyframe lists employed by the baseline methods. }
\label{fig:apartment-render}
\end{figure*}

\noindent \textbf{Implementation Details} The experiments were conducted on a single NVIDIA A100 GPU with 80 GB of VRAM. The keyframe interval was set to 20. The tracking and mapping iterations for the NeRF model are both set to 10, and the mapping iteration for the Gaussian map is 30. Parameters of the NeRF model and Gaussian primitives were optimized using the Adam optimizer, with the learning rate following the settings in \cite{muller2022instant} and \cite{kerbl20233d}. The initial pose is estimated purely using NeRF with a tracking iteration of 10. For mapping, NeRF undergoes 10 iterations, while the Gaussian map is optimized over 30 iterations for fine-grained reconstruction. The keyframe intervals for both NeRF and Gaussian models are 20 frames for the Replica \cite{straub2019replica} and TUM RGB-D \cite{sturm2012evaluating} datasets, and 25 frames for the ScanNet dataset \cite{dai2017scannet}. For NeRF, we employ a sampling ratio of 64 in our geometry-warping initialization. The density threshold $\tau_{\rm grid}$ for selecting transmission points is set at 0.001. In the ray-guided pruning process, the pruning threshold $\tau_{\rm prune}$ is set at 0.001 to optimize memory and computational resources. During map optimization, the regularization scale threshold $\tau_{\rm scale}$ is maintained at 1.0, where scales exceeding this threshold are penalized. The weighting hyperparameters $\lambda_{\rm c}$, $\lambda_{\rm d}$, $\lambda_{\rm ssim}$, and $\lambda_{\rm reg}$ are set at 0.8, 0.1, 0.2, and 0.001 respectively, promoting consistency across various metrics including color accuracy, depth precision, structural similarity, and regularization. \\

\noindent \textbf{Datasets and Metrics} We conduct comprehensive evaluations using both synthesized and real-world scenes from Replica~\cite{straub2019replica}, ScanNet~\cite{dai2017scannet}, and TUM RGB-D~\cite{sturm2012evaluating} datasets (supplementary material). In addition, we use four large-scale apartment scenes from the Replica dataset~\cite{straub2019replica}, with the largest scene containing up to two floors and eight rooms, presenting challenging indoor layouts with complex corridors and stairs. To evaluate tracking accuracy, we employ ATE RMSE (cm). For reconstruction results on training views, we adhere to PSNR, SSIM, and LPIPS to quantitatively evaluate rendering quality. To qualitatively compare mapping quality, we visualize the reconstruction outcomes from novel views. The running speed and computation usage are assessed using FPS and GPU consumption. Best results are shaded as \colorbox{db}{\textbf{first}}, \colorbox{mb}{\textbf{second}}, and \colorbox{lb}{\textbf{third}}. \\

\noindent \textbf{Baseline Methods} We compare DenseSplat with NeRF-based RGB-D SLAM systems, including Co-SLAM~\cite{wang2023co}, Point-SLAM~\cite{sandstrom2023point}, Loopy-SLAM~\cite{liso2024loopy}; as well as recent 3DGS-based systems such as SplaTAM~\cite{keetha2024splatam}, MonoGS~\cite{matsuki2024gaussian}, Gaussian-SLAM~\cite{yugay2023gaussian}, Photo-SLAM~\cite{huang2024photo}, and RTG-SLAM~\cite{peng2024rtg}. To ensure a fair comparison, all results are assessed based on the final global maps produced by the SLAM systems.

\begin{table}[t]
\centering
\caption{Evaluation of averaged tracking performance, system frame rate, memory consumption, and map size across the eight scenes of the Replica dataset~\cite{straub2019replica}. We separately present the systems that utilize ORB-SLAM~\cite{mur2017orb,campos2021orb} and frame-to-frame tracking.}
\footnotesize
\setlength{\tabcolsep}{0pt}
\begin{tabularx}{0.48\textwidth}{@{\hspace{1pt}}c@{\hspace{3pt}}|>{\hspace{3pt}}l@{\hspace{1pt}}|@{\hspace{1pt}}*{6}{>{\centering\arraybackslash}X}}
\hline
& Methods & ATE [cm]$\downarrow$ & Track. FPS$\uparrow$ & Map. FPS$\uparrow$ & System FPS$\uparrow$ & Memory [GB]$\downarrow$ & Size [MB]$\downarrow$ \\
\hline
\multirow{-0.9}{*}{\rotatebox[origin=c]{90}{ORB}} & Photo-SLAM~\cite{huang2024photo}~ & 0.59 & \cellcolor{mb}\textbf{41.64}  & \cellcolor{db}\textbf{30.36}  & \cellcolor{db}\textbf{20.71} & \cellcolor{lb}\textbf{8.02} &  \cellcolor{lb}\textbf{59} \\
& RTG-SLAM~\cite{peng2024rtg}~ & \cellcolor{lb}{0.49} & \cellcolor{db}\textbf{50.33}  & \cellcolor{lb}\textbf{20.09} & \cellcolor{mb}\textbf{17.30} & 10.18 & 71 \\
\hline
& Co-SLAM~\cite{wang2023co}~ & 1.12 & \cellcolor{lb}\textbf{10.2} & 10.0 & \cellcolor{lb}\textbf{9.26} & \cellcolor{db}\cellcolor{mb}\textbf{7.90} & \cellcolor{db}\textbf{7} \\
\multirow{-1.7}{*}{\rotatebox[origin=l]{90}{Frame-to-frame}} & Point-SLAM~\cite{sandstrom2023point}~ & {0.54} & 0.95 & 0.87  & 0.44 & 18.86 & 154  \\
& Loopy-SLAM~\cite{liso2024loopy}~ & \cellcolor{db}\textbf{0.29} & 0.95  & 0.87 & 0.43 & 18.91 & 177  \\
& SplaTAM~\cite{keetha2024splatam}~ & 0.55 & 0.86 & 0.51 & 0.51 & 11.27 & 331 \\
& MonoGS~\cite{matsuki2024gaussian}~ & 0.58 & 3.41 & 3.16 & 3.09 & 13.88 & \cellcolor{mb}\textbf{42} \\
& \textbf{Ours} &\cellcolor{mb}\textbf{0.33}  &7.96  & \cellcolor{mb}\textbf{23.35} & 7.67 & \cellcolor{db}\textbf{6.67} & 117 \\
\hline
\end{tabularx}
\label{tab:replica_fps}
\end{table}

\subsection {Evaluation of Tracking and Mapping}
\label{sec:eval-track-map}

\noindent \textbf{Evaluations on Replica Dataset} We quantitatively assessed the rendering quality of our method against NeRF-based and Gaussian-based SLAM systems on the Replica dataset~\cite{straub2019replica}, as presented in \Cref{tab:replica_psnr}. DenseSplat exhibits competitive rendering quality in single-room scale scenes, achieving state-of-the-art performance in some instances. Moreover, our method substantially outperforms baseline approaches in large-scale, multi-room environments, such as apartment scenes. Specifically, Gaussian SLAM baselines struggle due to ineffective bundle adjustment and cumbersome map representations, which do not adequately address the extensive camera movement and complex geometry characteristic of large-scale scenes. NeRF baselines provide robust tracking and efficient implicit scene representation but do not match our method in detailed reconstruction and rendering quality. DenseSplat integrates the strengths of both NeRF and Gaussian maps, utilizing the robust NeRF prior to stabilize scene representation. In addition, our submap strategy facilitates parallel computation of maps, significantly reducing memory consumption that often leads to system failures in Gaussian systems, such as SplaTAM~\cite{keetha2024splatam} in apartment\_0. The novel-view rendering results, shown in \Cref{fig:apartment-render}, qualitatively compare our reconstructed map with Gaussian-based systems. DenseSplat demonstrates superior mapping quality compared to baselines that suffer from significant scene drift and floaters.

To assess tracking accuracy and system efficiency, we compared the average ATE and frame rates as detailed in \Cref{tab:replica_fps}. DenseSplat delivers competitive tracking performance while achieving the highest mapping frame rate among frame-to-frame systems. Additionally, our map division strategy results in the lowest memory consumption during runtime compared to the baseline methods. \\

\begin{figure}[!tp]
\centering
\adjustbox{width=0.48\textwidth, scale={1}{1}}{
    \includegraphics{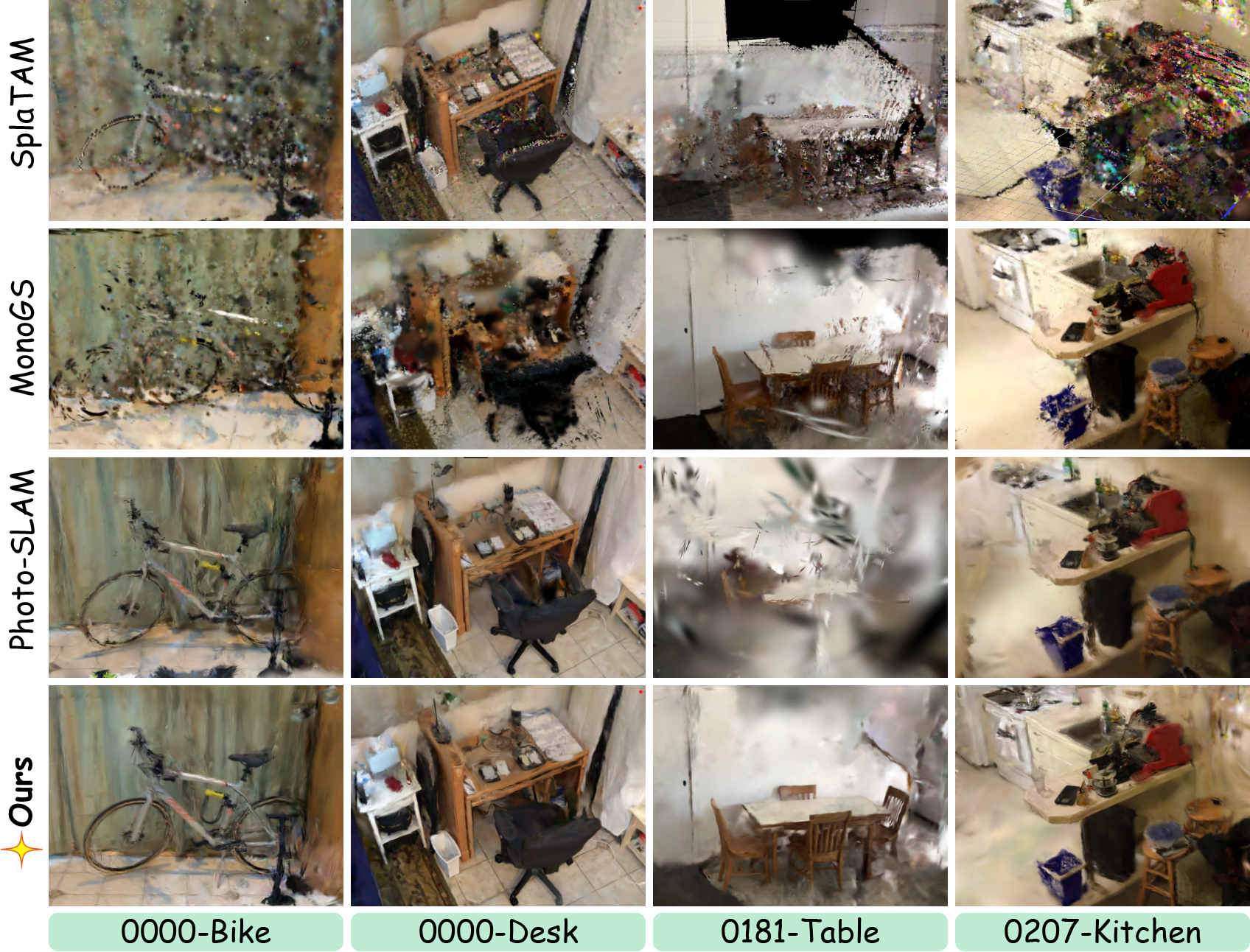}
}
\caption{\textbf{NVS} comparison on real-world ScanNet dataset \cite{dai2017scannet}. DenseSplat shows superior geometry accuracy and hole filling.}
\label{fig:scannet-render}
\end{figure}

\begin{figure*}[!tp]
\centering
\adjustbox{width=\textwidth, scale={1}{1}}{
    \includegraphics{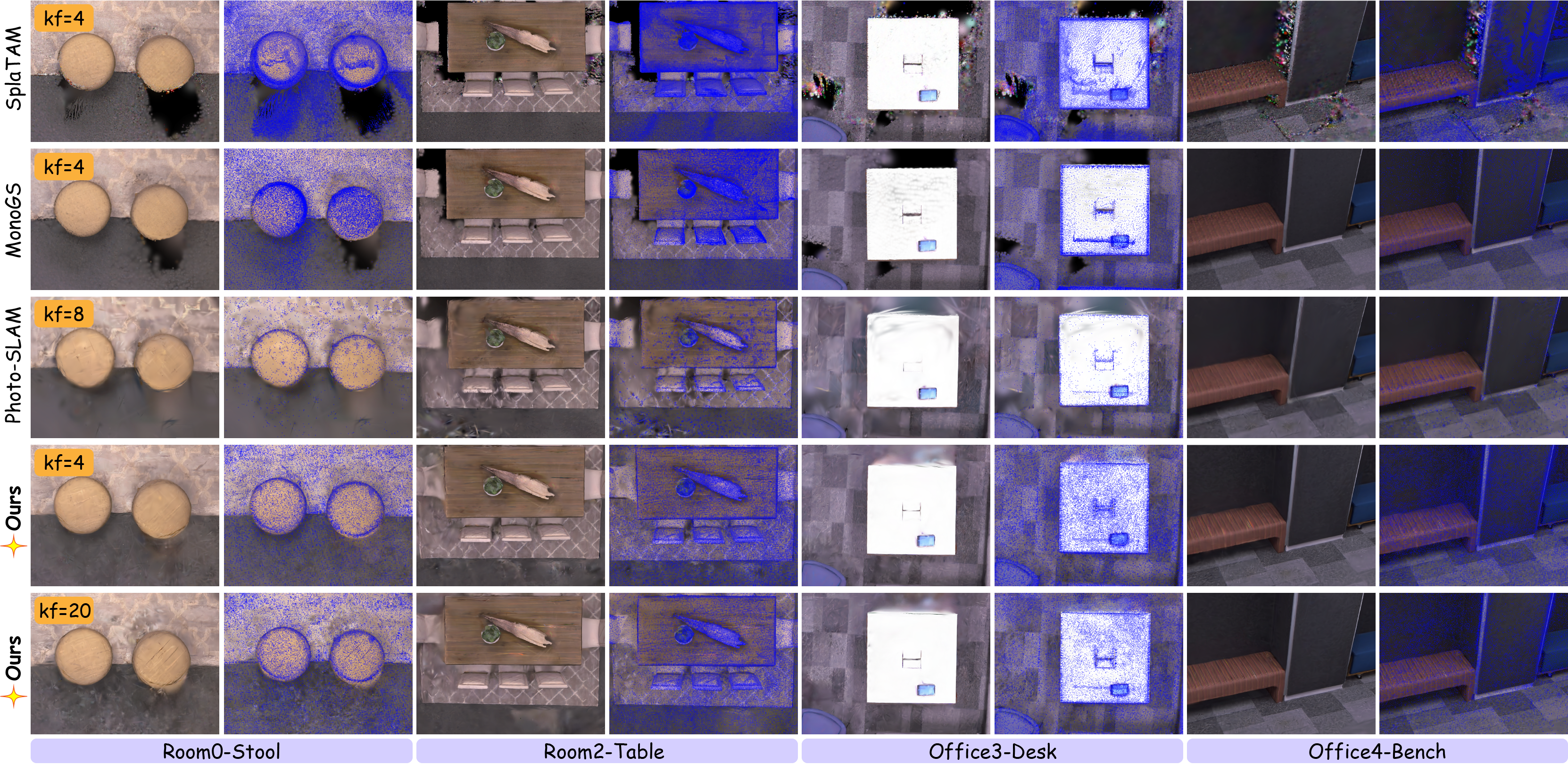}
}
\caption{Visualization of \textbf{NVS} of the Replica dataset \cite{straub2019replica}. The rendered views are shown along with plots for the center of primitives, depicted as blue dots of the same size across all methods. For our method, we explicitly present the NVS at keyframe intervals (kf) at 4 and 20. Looking at objects like floors and tables, DenseSplat demonstrates more evenly distributed primitives compared to baseline methods.}
\label{fig:scene-densification}
\end{figure*}

\noindent \textbf{Evaluations on ScanNet Dataset} We present quantitative evaluations of our method alongside the baseline approaches using 6 scenes from the ScanNet dataset \cite{dai2017scannet}. As shown in \Cref{tab:sup_scannet_psnr}, we evaluate rendering quality by comparing PSNR, SSIM, and LPIPS metrics across training views. In real-world environments, challenges on noisy input such as depth map errors and blurred images introduce significant difficulties for map initialization and optimization. Additionally, recent Gaussian-based methods like SplaTAM \cite{keetha2024splatam} and MonoGS \cite{matsuki2024gaussian}, which do not incorporate loop-closure, experience considerable loop-induced map drift that remarkably reduces reconstruction quality. In contrast, our method utilizes the robust NeRF prior and integrates both the bundle adjustment (BA) and subsequent BA-induced map refinement modules, achieving state-of-the-art performance. \Cref{tab:sup_scannet_ate} also presents the tracking performance using ATE RMSE (cm) on the ScanNet dataset \cite{dai2017scannet}. Our method demonstrates superior tracking accuracy compared to Gaussian baseline methods \cite{keetha2024splatam, matsuki2024gaussian} that do not incorporate the BA process. When comparing with Photo-SLAM \cite{huang2024photo} and RTG-SLAM \cite{peng2024rtg}, which incorporate point-based tracking systems derived from ORB-SLAM \cite{mur2017orb, campos2021orb}, DenseSplat achieves lower ATE errors, benefiting from its robust tracking based on the NeRF model.

\begin{table}[!tp]
    \centering
    \caption{Quantitative comparison of our method and the baselines in training view rendering on the ScanNet dataset \cite{dai2017scannet}. Our method demonstrates \colorbox{db}{SOTA} performances.}
    \setlength{\tabcolsep}{0pt} 
    \footnotesize 
    \begin{tabularx}{0.48\textwidth}{@{}l@{\hspace{0pt}} | >{\hspace{2pt}}c@{\hspace{1pt}} *{8}{>{\centering\arraybackslash}X}@{}}
        \Xhline{2\arrayrulewidth}
        \textbf{Methods} & \textbf{Metrics} & 0000 & 0059 & 0106 & 0169 & 0181 & 0207 & \textbf{Avg.} \\
        \hline
        & PSNR$\uparrow$ & 17.81 & 19.60 & 19.23 & 20.55 & 16.76 & 17.95 & 18.65 \\
        SplaTAM~\cite{keetha2024splatam}~ & SSIM$\uparrow$ & 0.602 & \cellcolor{lb}\textbf{0.796} & 0.741 & \cellcolor{lb}\textbf{0.785} & 0.683 & 0.705 & 0.719 \\
        & LPIPS$\downarrow$ & 0.467 & \cellcolor{lb}\textbf{0.290} & 0.322 & \cellcolor{lb}\textbf{0.260} & 0.420 & 0.346 & \cellcolor{lb}\textbf{0.351} \\
        \hline
        & PSNR$\uparrow$ & 18.62 & 15.56 & 14.97 & 18.07 & 14.98 & 18.52 & 16.79 \\
        RTG-SLAM~\cite{peng2024rtg}~ & SSIM$\uparrow$ & 0.756 & 0.682 & 0.726 & 0.772 & 0.750 & \cellcolor{lb}\textbf{0.773} & 0.743 \\
        & LPIPS$\downarrow$ & 0.468 & 0.531 & 0.480 & 0.451 & 0.492 & 0.459 & 0.480 \\
        \hline
        & PSNR$\uparrow$ & 19.74 & \cellcolor{lb}\textbf{20.01} & 18.70 & 20.23 & 14.45 & 19.96 & 18.85 \\
        Photo-SLAM~\cite{huang2024photo}~ & SSIM$\uparrow$ & \cellcolor{lb}\textbf{0.761} & \cellcolor{mb}\textbf{0.799} & \cellcolor{lb}\textbf{0.760} & 0.781 & 0.698 & 0.765 & \cellcolor{lb}\textbf{0.761} \\
        & LPIPS$\downarrow$ & \cellcolor{lb}\textbf{0.412} & \cellcolor{mb}\textbf{0.284} & \cellcolor{lb}\textbf{0.297} & 0.288 & 0.521 & \cellcolor{lb}\textbf{0.315} & 0.353 \\
        \hline
        & PSNR$\uparrow$ & \cellcolor{mb}\textbf{20.72} & \cellcolor{mb}\textbf{20.06} & \cellcolor{mb}\textbf{21.20} & \cellcolor{mb}\textbf{22.32} & \cellcolor{mb}\textbf{22.29} & \cellcolor{mb}\textbf{22.80} & \cellcolor{mb}\textbf{21.56} \\
        Gauss-SLAM~\cite{yugay2023gaussian}~ & SSIM$\uparrow$ & 0.702 & 0.728 & \cellcolor{lb}\textbf{0.785} & 0.764 & 0.774 & 0.769 & 0.754 \\
        & LPIPS$\downarrow$ & 0.568 & 0.496 & 0.460 & 0.458 & 0.544 & 0.505 & 0.505 \\
        \hline
        & PSNR$\uparrow$ & \cellcolor{lb}\textbf{19.76} & 19.25 & \cellcolor{lb}\textbf{20.18} & \cellcolor{lb}\textbf{20.57} & \cellcolor{lb}\textbf{20.25} & \cellcolor{lb}\textbf{20.62} & \cellcolor{lb}\textbf{20.11} \\
        MonoGS~\cite{matsuki2024gaussian}~ & SSIM$\uparrow$ & \cellcolor{mb}\textbf{0.772} & 0.767 & \cellcolor{mb}\textbf{0.785} & \cellcolor{mb}\textbf{0.790} & \cellcolor{lb}\textbf{0.788} & \cellcolor{mb}\textbf{0.798} & \cellcolor{mb}\textbf{0.783} \\
        & LPIPS$\downarrow$ & \cellcolor{mb}\textbf{0.387} & 0.289 & \cellcolor{mb}\textbf{0.272} & \cellcolor{mb}\textbf{0.256} & \cellcolor{mb}\textbf{0.282} & \cellcolor{mb}\textbf{0.295} & \cellcolor{mb}\textbf{0.297} \\
        \hline
        \noalign{\vskip 0.2pt}
        & PSNR$\uparrow$ & \cellcolor{db}\textbf{25.31} & \cellcolor{db}\textbf{24.74} &
        \cellcolor{db}\textbf{25.33} &
        \cellcolor{db}\textbf{24.98} &
        \cellcolor{db}\textbf{23.61} &
        \cellcolor{db}\textbf{25.29} &
        \cellcolor{db}\textbf{24.88} \\
        \textbf{Ours} & SSIM$\uparrow$ &
        \cellcolor{db}\textbf{0.832} &
        \cellcolor{db}\textbf{0.875} &
        \cellcolor{db}\textbf{0.866} &
        \cellcolor{db}\textbf{0.862} &
        \cellcolor{db}\textbf{0.845} &
        \cellcolor{db}\textbf{0.847} &
        \cellcolor{db}\textbf{0.855} \\
        & LPIPS$\downarrow$ &
        \cellcolor{db}\textbf{0.206} &
        \cellcolor{db}\textbf{0.211} &
        \cellcolor{db}\textbf{0.195} &
        \cellcolor{db}\textbf{0.215} &
        \cellcolor{db}\textbf{0.250} &
        \cellcolor{db}\textbf{0.198} &
        \cellcolor{db}\textbf{0.212} \\
        \Xhline{2\arrayrulewidth}
     \end{tabularx}
    \label{tab:sup_scannet_psnr}
\end{table}

\begin{table}[!tp]
\centering
\caption{The performance of ATE RMSE (cm) on 6 scenes from the ScanNet~\cite{dai2017scannet} dataset. For our method, we provide results both with and without Bundle Adjustment (BA).}
\footnotesize
\setlength{\tabcolsep}{0pt}
\begin{tabularx}{0.48\textwidth}{@{}@{\hspace{1pt}}c@{\hspace{3pt}}|>{\hspace{2pt}}l@{\hspace{1pt}}|@{\hspace{1pt}}*{7}{>{\centering\arraybackslash}X}}
\hline
& Methods & 0000 & 0059 & 0106 & 0169 & 0181 & 0207 & \textbf{Avg.} \\
\hline
\multirow{-1}{*}{\rotatebox[origin=c]{90}{ORB}} & Photo-SLAM~\cite{huang2024photo}~ & 7.62 & 7.94 & 9.36 & 10.01 &22.97  & 7.23 &10.85 \\
& RTG-SLAM~\cite{peng2024rtg}~ & 8.04 & \cellcolor{mb}\textbf{6.82} & 9.22 & 10.15 &24.36  & 9.25 &11.31  \\
\hline
& ESLAM~\cite{johari2023eslam}~ & 7.54 & 8.52 & \cellcolor{db}\textbf{7.39} & 8.17 & \cellcolor{mb}\textbf{9.13} & \cellcolor{db}\textbf{5.61} & \cellcolor{mb}\textbf{7.73} \\
& Co-SLAM~\cite{wang2023co}~ & \cellcolor{mb}\textbf{7.13} & 11.14 & 9.36 & \cellcolor{db}\textbf{5.90} & 11.81 & \cellcolor{mb}\textbf{7.14} & 8.75 \\
\multirow{-2}{*}{\rotatebox[origin=l]{90}{Frame-to-frame}} & Loopy-SLAM~\cite{liso2024loopy} & \cellcolor{db}\textbf{4.28} & \cellcolor{lb}\textbf{7.59} & 8.37 & 7.56 & 10.68 & 7.95 & \cellcolor{db}\textbf{7.70} \\
& SplaTAM~\cite{keetha2024splatam}~ & 12.83 & 10.10 & 17.72 & 12.08 & 11.10 & 7.47 & 11.88 \\
& MonoGS~\cite{matsuki2024gaussian}~ & 15.94 & \cellcolor{db}\textbf{6.41} & 19.44 & 10.44 & 12.23 & 10.46 & 12.48 \\
& Ours (w/o BA) & 7.94 & 10.85 &\cellcolor{lb}\textbf{7.62} & \cellcolor{lb}\textbf{6.91} & \cellcolor{lb}\textbf{9.20} & 8.33 &8.48   \\
& \textbf{Ours} & \cellcolor{lb}\textbf{7.25} & 9.60 & \cellcolor{mb}\textbf{7.44} & \cellcolor{mb}\textbf{6.58} & \cellcolor{db}\textbf{8.71} & \cellcolor{lb}\textbf{7.21} & \cellcolor{lb}\textbf{7.80} \\
\hline
\end{tabularx}
\label{tab:sup_scannet_ate}
\end{table}

\Cref{fig:scannet-render} shows the novel-view synthesis of representative scenes from the ScanNet dataset \cite{dai2017scannet}. DenseSplat offers superior geometric accuracy, for instance, accurately capturing the details of the bicycle in scene\_0000, which often suffers from scene drift due to trajectory loops. Compared to Photo-SLAM \cite{huang2024photo}, our method demonstrates comparable tracking performance while providing more fine-grained and complete map reconstruction.\\

\begin{figure*}[!tp]
\centering
\includegraphics[width=\textwidth]{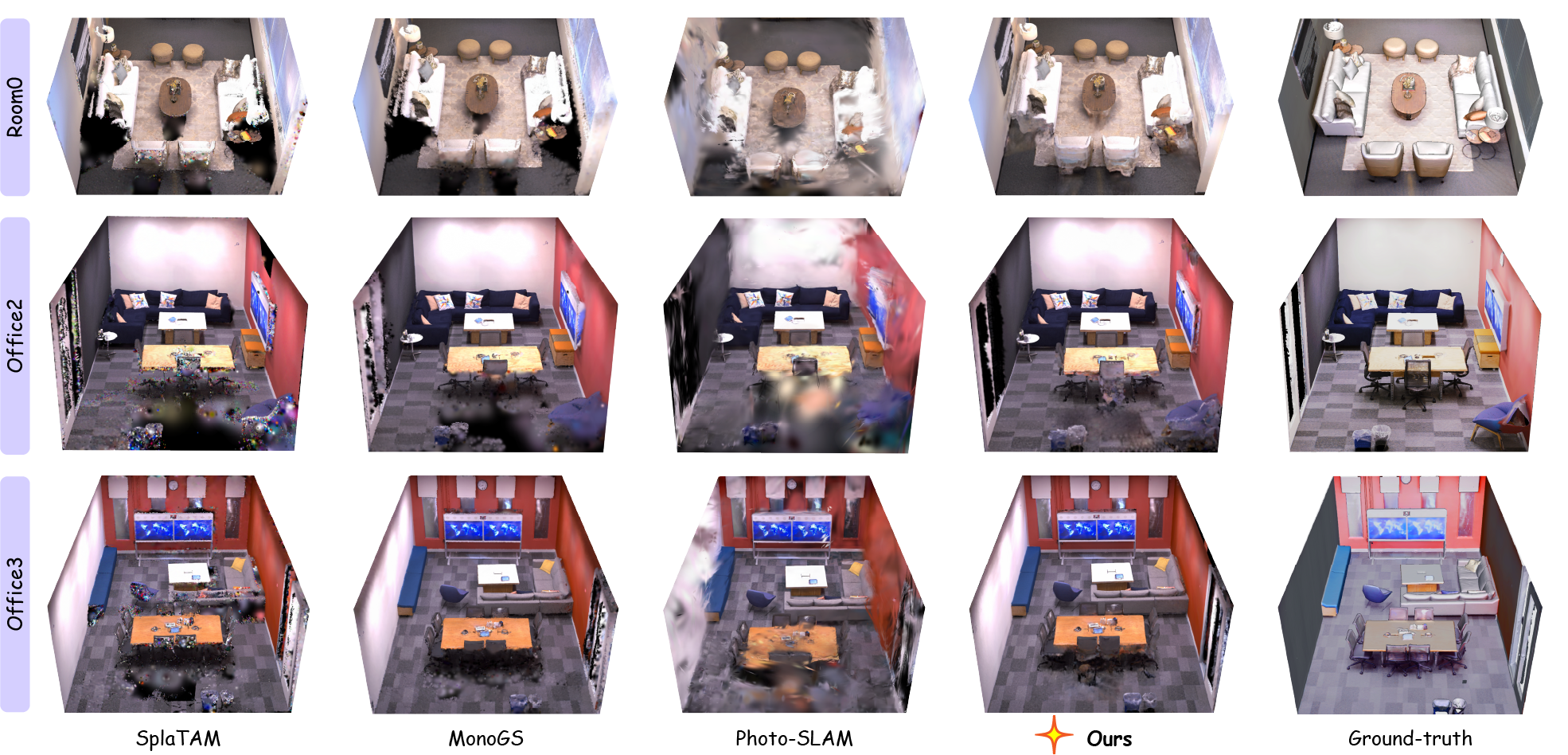}
\caption{Visualization of hole-filling in novel-view synthesis for the scene Room\_0, Office\_2, and Office\_3 from the Replica dataset \cite{straub2019replica}. Employing the NeRF prior, our method effectively interpolates the unobserved geometry obscured by obstacles in the room, contrasting with Gaussian-based approaches that exhibit significant gaps, drastically compromising the completeness of the reconstruction.}
\label{sup-fig:success-case}
\end{figure*}

\begin{table}[!tp]
\centering
\caption{The performance of ATE RMSE (cm) on the TUM RGB-D dataset \cite{sturm2012evaluating}. For our method, we provide results both with and without Bundle Adjustment (BA).}
\footnotesize
\setlength{\tabcolsep}{0pt}
\begin{tabularx}{0.48\textwidth}{@{}@{\hspace{1pt}}c@{\hspace{3pt}}|>{\hspace{2pt}}l@{\hspace{1pt}}|@{\hspace{1pt}}*{4}{>{\centering\arraybackslash}X}}
\hline
& Methods & fr1\_desk & fr2\_xyz & fr3\_office & \textbf{Avg.} \\
\hline
\multirow{-1}{*}{\rotatebox[origin=c]{90}{ORB}} & Photo-SLAM~\cite{huang2024photo}~ & 2.61 & \cellcolor{db}\textbf{0.35} & \cellcolor{db}\textbf{1.00} & \cellcolor{mb}\textbf{1.32} \\
& RTG-SLAM~\cite{peng2024rtg}~ & \cellcolor{db}\textbf{1.66} & \cellcolor{mb}\textbf{0.38} & \cellcolor{mb}\textbf{1.13} & \cellcolor{db}\textbf{1.06} \\
\hline
& ESLAM~\cite{johari2023eslam}~ & 2.30 & 1.10 & 2.40 & 2.00 \\
& Co-SLAM~\cite{wang2023co}~ & 2.70 & 1.90 & 2.60 & 2.40 \\
\multirow{-2}{*}{\rotatebox[origin=l]{90}{Frame-to-frame}} & Loopy-SLAM~\cite{liso2024loopy} & 3.79 & 1.62 & 3.41 & 2.94 \\
& SplaTAM~\cite{keetha2024splatam}~ & 3.35 & 1.24 & 5.16 & 3.25 \\
& MonoGS~\cite{matsuki2024gaussian}~ & \cellcolor{mb}\textbf{1.52} & 1.58 & \cellcolor{lb}\textbf{1.65} & \cellcolor{lb}\textbf{1.58} \\
& Ours (w/o BA) &\cellcolor{lb}\textbf{2.02}  &0.97  &2.48  &1.82  \\
& \textbf{Ours} &\cellcolor{lb}\textbf{2.02}  &\cellcolor{lb}\textbf{0.86}  &1.86  & \cellcolor{lb}\textbf{1.58} \\
\hline
\end{tabularx}
\label{tab:sup_tum_ate}
\end{table}

\noindent \textbf{Evaluations on TUM RGB-D Dataset} We also conducted experiments on the TUM RGB-D dataset \cite{sturm2012evaluating} and compared the ATE RMSE (cm) with both feature-based and frame-to-frame tracking systems in \Cref{tab:sup_tum_ate}. Our system achieves performance comparable to current state-of-the-art methods.

\subsection{Evaluation of Scene Densification}
\label{sec:result-scene-densification}

\Cref{fig:scene-densification} illustrates the novel-view synthesis of DenseSplat and Gaussian-based SLAM systems on the Replica dataset~\cite{straub2019replica}. The obstructed views typical of indoor environments pose remarkable challenges for Gaussian SLAM systems. For instance, SplaTAM~\cite{keetha2024splatam} and MonoGS \cite{matsuki2024gaussian} exhibit notable deficiencies with holes in their reconstructed Gaussian maps. Although Photo-SLAM~\cite{huang2024photo} mitigates these gaps using a pyramid feature extraction strategy, it results in sparser map representations, which diminish rendering quality. In contrast, DenseSplat employs the robust NeRF prior for Gaussian initialization, effectively filling these gaps by interpolating unobserved geometry. Our method also ensures a more uniform distribution of Gaussian primitives across the surface of the objects using our point sampling strategy, leading to better geometric alignment. Furthermore, \Cref{fig:scene-densification} also compares the rendering results of our method at a dense keyframe interval of 4, typical of current Gaussian SLAM systems, with a much sparser interval of 20. Despite the map being slightly less populated, DenseSplat continues to demonstrate robust reconstruction and interpolation capabilities. This effectiveness in sparse conditions significantly enhances real-time processing and, more importantly, relaxes the stringent requirement for dense multi-view observations previously essential in Gaussian SLAM systems. \Cref{sup-fig:success-case} provides a broader view of the Replica scenes, showcasing the superior map completeness of our method compared to baseline approaches. The latter often display significant gaps due to common occlusions encountered in indoor environments.

\begin{table}[!tp]
    \setlength{\tabcolsep}{0pt} 
    \footnotesize 
    \centering
    \caption{The mesh evaluation of our method and the Gaussian SLAM baselines on the Replica dataset \cite{straub2019replica}. The results are averaged over 8 scenes.}
    \begin{tabularx}{0.48\textwidth}{@{\hspace{1pt}}c@{\hspace{3pt}}|>{\hspace{3pt}}l@{\hspace{1pt}}|@{\hspace{1pt}}*{3}{>{\centering\arraybackslash}X}}
        \hline
        & Methods & Accuracy$\downarrow$ & Completion$\downarrow$ & Comp. Ratio $\uparrow$ \\
        \hline
        \multirow{-0.5}{*}{\rotatebox[origin=c]{90}{NeRF}} & NICE-SLAM \cite{zhu2022nice} & {2.85} & {3.02} & {89.34} \\
        & Co-SLAM~\cite{wang2023co} & \cellcolor{mb}\textbf{2.10} & \cellcolor{lb}\textbf{2.08} & \cellcolor{lb}\textbf{93.44} \\
        & ESLAM~\cite{johari2023eslam} & \cellcolor{db}\textbf{0.97} & \cellcolor{db}\textbf{1.05} & \cellcolor{db}\textbf{98.60} \\
        \hline
        \multirow{-0.1}{*}{\rotatebox[origin=c]{90}{3DGS}} & SplaTAM~\cite{keetha2024splatam} & 2.74 & 4.02 & 84.89 \\
        & MonoGS~\cite{matsuki2024gaussian} & 3.16 & 4.45 & 81.52 \\
        & Photo-SLAM~\cite{huang2024photo} & 2.53 & 3.75 & 85.67 \\
        & \textbf{Ours}  & \cellcolor{lb}\textbf{2.18} & \cellcolor{mb}\textbf{2.01} & \cellcolor{mb}\textbf{94.64} \\ 
        \hline
    \end{tabularx}
    \label{tab:mesh_accuracy}
\end{table}

Moreover, to quantitatively evaluate the reconstruction completeness, we transform Gaussian maps into meshes using TSDF-fusion \cite{curless1996volumetric} and compare them with ground-truth meshes.
As demonstrated in \Cref{tab:mesh_accuracy}, DenseSplat significantly outperforms recent Gaussian SLAM systems in accuracy and scene completion ratio, which often struggle with gaps that result in underrepresented scenes. \\

\label{sec:results_densification}

\begin{figure*}[!tp]
\centering
\includegraphics[width=\textwidth]{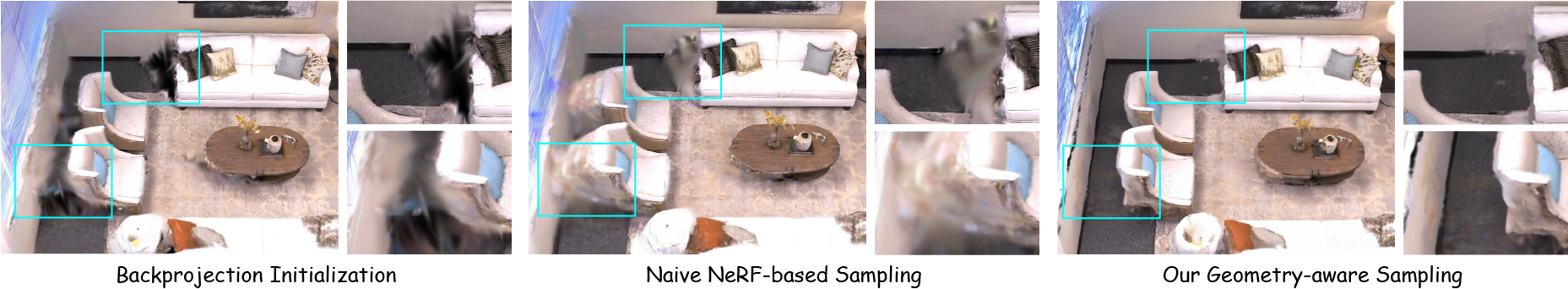}
\caption{Visualization of the ablation study on Gaussian primitive initialization approaches for Room0 of the Replica dataset \cite{straub2019replica}.
For direct backprojection, we backproject the RGB-D frames and downsample by $\times 1$ to initialize primitives.
For native NeRF-based sampling, we uniformly sample NeRF using a grid size of 0.01. For our geometry‐aware sampling, we use $\tau_{\rm grid}=0.001$ and the same sampling grid size for initialization. Our strategy achieves superior geometric accuracy compared to direct backprojection, which exhibits substantial holes, and the naive NeRF‐based method, which produces many redundant floating primitives.}
\label{fig:ablation_gs_init}
\end{figure*}

\section{Ablation Study}

\begin{table}[!tp]
    \setlength{\tabcolsep}{0pt} 
    \footnotesize 
    \centering
    \caption{Ablation study of each component of DenseSplat on Room0 of the Replica dataset \cite{straub2019replica}.}
    \begin{tabularx}{0.48\textwidth}{l@{\hspace{1pt}}|@{\hspace{1pt}}*{5}{>{\centering\arraybackslash}X}}
        \hline
        Ablation Method & PSNR [dB]$\uparrow$ & Render. FPS$\uparrow$ & Map. FPS$\uparrow$ & Num. of $\mathcal{G}$~[K]$\downarrow$ & Memory [GB]$\downarrow$ \\
        \hline
        w/o pruning &32.01  &19.96  &6.94  &97.98  & 10.84    \\
        w/o BA loss  &\cellcolor{mb}\textbf{36.66}  & 23.87  &\cellcolor{mb}\textbf{8.76}  &\cellcolor{lb}\textbf{40.98}  &\cellcolor{db}\textbf{6.09}    \\
        w/o multi-scale GS &33.64  &20.09  &7.99  &55.91  &\cellcolor{lb}\textbf{9.67}    \\
        w/o submap &\cellcolor{lb}\textbf{36.60} &\cellcolor{lb}\textbf{22.46}  &7.01  &\cellcolor{mb}\textbf{39.40}  &9.86    \\
        \hline
        \textbf{Ours}  &\cellcolor{db}\textbf{37.11} & \cellcolor{mb}\textbf{23.99} &\cellcolor{lb}\textbf{8.08}  &\cellcolor{db}\textbf{37.52}  &\cellcolor{mb}\textbf{6.68} \\ 
        \hline
    \end{tabularx}
    \label{tab:ablation_overall}
\end{table}

In this section, we conduct extensive ablation studies on each hyperparameter and core component of our system in terms of rendering quality, scene accuracy, and system efficiency. The overall ablation analysis is presented in \Cref{tab:ablation_overall}. By taking the tradeoff of rendering and mapping efficiency, our method capitalizes on the benefits of high-quality rendering using the NeRF model. \\

\begin{table}[!tp]
    \setlength{\tabcolsep}{0pt}
    \footnotesize
    \centering
    \caption{Ablation study on the initialization of primitives $\mathcal{G}$, conducted on Room0 from the Replica dataset \cite{straub2019replica}. For the backprojection-based initialization methods shown in the middle rows, we directly backproject pixels from the RGB-D input frame, using downsampling ratios of $\times1$, $\times8$, $\times16$.}
    \begin{tabularx}{0.48\textwidth}{l@{\hspace{1pt}}|@{\hspace{1pt}}*{5}{>{\centering\arraybackslash}X}}
        \hline
        Initialization Method & PSNR [dB]$\uparrow$ & Render. FPS$\uparrow$ & Map. FPS$\uparrow$ & Num. $\mathcal{G}$~[K]$\downarrow$ & Memory [GB]$\downarrow$ \\
        \hline
        Random initialized $\mathcal{G}$ & 18.19  & \cellcolor{db}\textbf{117.47} & \cellcolor{lb}\textbf{10.12} & \cellcolor{mb}\textbf{190.64} & \cellcolor{mb}\textbf{17.13}    \\
        \hline
        Backprojected $\mathcal{G}$ (down ${\times 1}$) & \cellcolor{mb}\textbf{36.26} & 39.96 & \textbf{8.33}  & 512.04 & 23.49    \\
        Backprojected $\mathcal{G}$ (down ${\times 8}$) & 33.27 & \cellcolor{lb}\textbf{57.13} & \cellcolor{mb}\textbf{11.56}  & 339.17 & 20.39 \\
        Backprojected $\mathcal{G}$ (down ${\times 16}$) & 31.69  & \cellcolor{mb}\textbf{92.17} &\cellcolor{db}\textbf{15.77} & \cellcolor{lb}\textbf{205.82} & \cellcolor{lb}\textbf{18.96} \\
        \hline
        Naive NeRF sampling & \cellcolor{lb}\textbf{34.12} & 16.54 & 6.64 & 289.36 & 21.31 \\
        Geometry-aware sampling (\textbf{ours}) &\cellcolor{db}\textbf{37.11} & \textbf{23.99}  &\textbf{8.08} & \cellcolor{db}\textbf{37.52} & \cellcolor{db}\textbf{6.68} \\
        \hline
    \end{tabularx}
    \label{tab:ablation_nerf_init}
\end{table}

\noindent \textbf{Ablation of NeRF-based Initialization} To evaluate the effectiveness of our geometry-aware sampling strategy, \Cref{tab:ablation_nerf_init} presents an ablation study on different primitive initialization strategies. We pay particular attention to the method of direct backprojection from RGB-D streams, which is commonly applied in previous Gaussian-based SLAM systems \cite{keetha2024splatam,matsuki2024gaussian,yugay2023gaussian}. Specifically, the baseline approaches include random primitive initialization, direct backprojection of RGB-D frames downsampled by $\times 1$, $\times 8$, and $\times 16$, and a naive NeRF-based sampling approach that uniformly samples NeRF without removing redundant points. The ablation results shows that, by actively selecting surface transition points, our geometry-aware point sampling strategy achieves optimal rendering quality while consuming the least memory.

Furthermore, \Cref{fig:ablation_gs_init} shows novel-view synthesis results for Room0 using direct backprojection, naive NeRF-based sampling, and our adaptive strategy. Direct backprojection leaves substantial holes due to insufficient or occluded views, while naive NeRF-based sampling introduces floating primitives due to the lack of filtering. Our method delivers a complete reconstruction, effectively filling gaps and maintaining geometric accuracy. \\

\noindent \textbf{Ablation of Thresholding Hyperparameters} We investigate three thresholding hyperparameters in \Cref{tab:ablation_threshold}: $\tau_{\rm grid}$ for NeRF sampling, $\tau_{\rm prune}$ for primitive pruning, and $\tau_{\rm scale}$ for scale regularization. Setting $\tau_{\rm grid}=0$ is equivalent to naive NeRF sampling. We employ a small sampling threshold $\tau_{\rm grid}=0.001$ to effectively remove erroneous sampling points from the NeRF prior, improving geometry accuracy and system efficiency. The parameter $\tau_{\rm prune}$ controls the removal of primitives with negligible radiance contribution; a small threshold brings certain improvement in quality and efficiency, while a larger value remarkably accelerates rendering but degrades performance by pruning too many primitives. The parameter $\tau_{\rm scale}$ penalizes excessively large Gaussian primitives to suppress view-dependent artifacts, especially in sparse-view settings. Smaller thresholds encourage finer primitives at the cost of increased memory usage, whereas $\tau_{\rm scale} = 1$ offers a balanced trade-off between primitive granularity and overall efficiency. \\

\begin{table}[!tp]
    \setlength{\tabcolsep}{0pt}
    \footnotesize
    \centering
    \caption{Ablation study of thresholding hyperparameters, conducted on Room0 from the Replica dataset \cite{straub2019replica}. Note that $\tau_{\rm grid}=0$ is equivalent to naive NeRF-based initialization.}
    \begin{tabularx}{0.48\textwidth}{l@{\hspace{1pt}}|@{\hspace{1pt}}*{5}{>{\centering\arraybackslash}X}}
        \hline
        Sampling Strategy & PSNR [dB]$\uparrow$ & Render. FPS$\uparrow$ & Map. FPS$\uparrow$ & Num. $\mathcal{G}$~[K]$\downarrow$ & Memory [GB]$\downarrow$ \\
        \hline
        $\tau_{\rm grid}=0$ & 34.12 & 16.54 & 6.64 & 289.36 & 21.31 \\
        $\tau_{\rm grid}=0.001$ (\textbf{ours}) & \cellcolor{db}\textbf{37.11} & \cellcolor{db}\textbf{23.99} & \cellcolor{db}\textbf{8.08} & \cellcolor{db}\textbf{37.52} & \cellcolor{db}\textbf{6.68} \\
        $\tau_{\rm grid}=0.005$ & {36.58} & {23.07} & {7.61} & {55.97} & {7.74} \\
        $\tau_{\rm grid}=0.01$ & {35.56} & {20.15} & {6.70} & {78.63} & {9.87} \\
        \hline
        $\tau_{\rm prune}=0$ & {31.97} & {20.12} & {6.21} & {299.36} & {21.67} \\
        $\tau_{\rm prune}=0.001$ (\textbf{ours}) & \cellcolor{db}\textbf{37.11} & {23.99} & {8.08} & {37.52} &
        {6.68} \\
        $\tau_{\rm prune}=0.01$ & {35.19} & \cellcolor{db}\textbf{31.91} & \cellcolor{db}\textbf{11.54}  & \cellcolor{db}\textbf{21.23} & \cellcolor{db}\textbf{5.08} \\
        \hline
        $\tau_{\rm scale}=0.01$ &{35.15}  &17.77  &5.34  &160.77  & 11.25 \\
        $\tau_{\rm scale}=0.1$ &\cellcolor{db}\textbf{37.20}  &20.94  &6.89  &70.61  & 9.34 \\
        $\tau_{\rm scale}=1$ (\textbf{ours}) & {37.11} & \cellcolor{db}\textbf{23.99} & \cellcolor{db}\textbf{8.08} & \cellcolor{db}\textbf{37.52} & \cellcolor{db}\textbf{6.68} \\
        \hline
    \end{tabularx}
    \label{tab:ablation_threshold}
\end{table}     

\begin{figure}[!tp]
\centering
\includegraphics[width=0.48\textwidth]{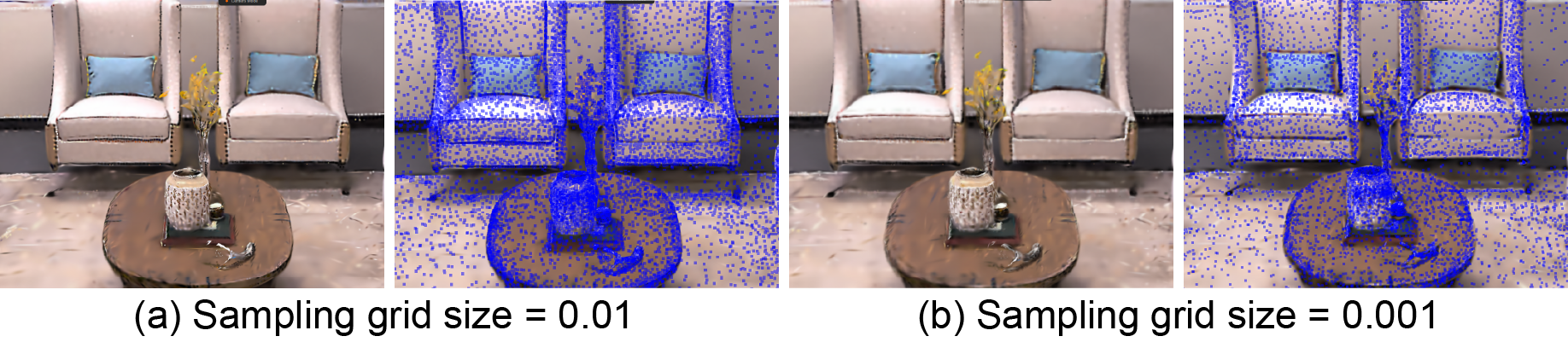}
\caption{Ablation study on sampling grid size, showing rendered images of Room0 from the Replica dataset \cite{straub2019replica}, with primitive's centers $\mu$ shaded in blue. (a) Sampling size of 0.01 (b) Sampling size of 0.001.}
\label{fig:abaltion-granuarity}
\end{figure}

\begin{table}[!tp]
    \setlength{\tabcolsep}{0pt}
    \footnotesize
    \centering
    \caption{Ablation study of sampling grid size during NeRF sampling, conducted on Room0 from the Replica dataset \cite{straub2019replica}.}
    \begin{tabularx}{0.48\textwidth}{l@{\hspace{1pt}}|@{\hspace{1pt}}*{5}{>{\centering\arraybackslash}X}}
        \hline
        Hyperparameters & PSNR [dB]$\uparrow$ & Render. FPS$\uparrow$ & Map. FPS$\uparrow$ & Num. $\mathcal{G}$~[K]$\downarrow$ & Memory [GB]$\downarrow$ \\
        \hline
        $\text{grid\_size}=0.001$ &{36.28}  &21.15  & 6.93  &69.66  & 9.27 \\
        $\text{grid\_size}=0.01$ (\textbf{ours}) & \cellcolor{db}\textbf{37.11} & {23.99} & {8.08} & {37.52} & 6.68 \\
        $\text{grid\_size}=0.1$ &{35.44}  & \cellcolor{db}\textbf{27.26}  & \cellcolor{db}\textbf{9.88}  &\cellcolor{db}\textbf{25.91}  & \cellcolor{db}\textbf{5.37} \\
        \hline
    \end{tabularx}
    \label{tab:ablation_grid_size}
\end{table}

\noindent \textbf{Ablation of Sampling Size} Although precise control over the number of Gaussian primitives is difficult due to the densification process \cite{kerbl20233d}, adjusting the sampling grid size of the NeRF model can roughly control the granularity of the scene representation. We perform an ablation study on the sampling grid size, as shown in \Cref{tab:ablation_grid_size}. The corresponding visualization in \Cref{fig:abaltion-granuarity} shows that, compared to our choice of 1 cm, increasing the sampling grid size by a factor of 10 produces a sparser primitive distribution and reduces rendering quality to some extent. \\

\begin{figure}[ht]
\centering
\includegraphics[width=0.48\textwidth]{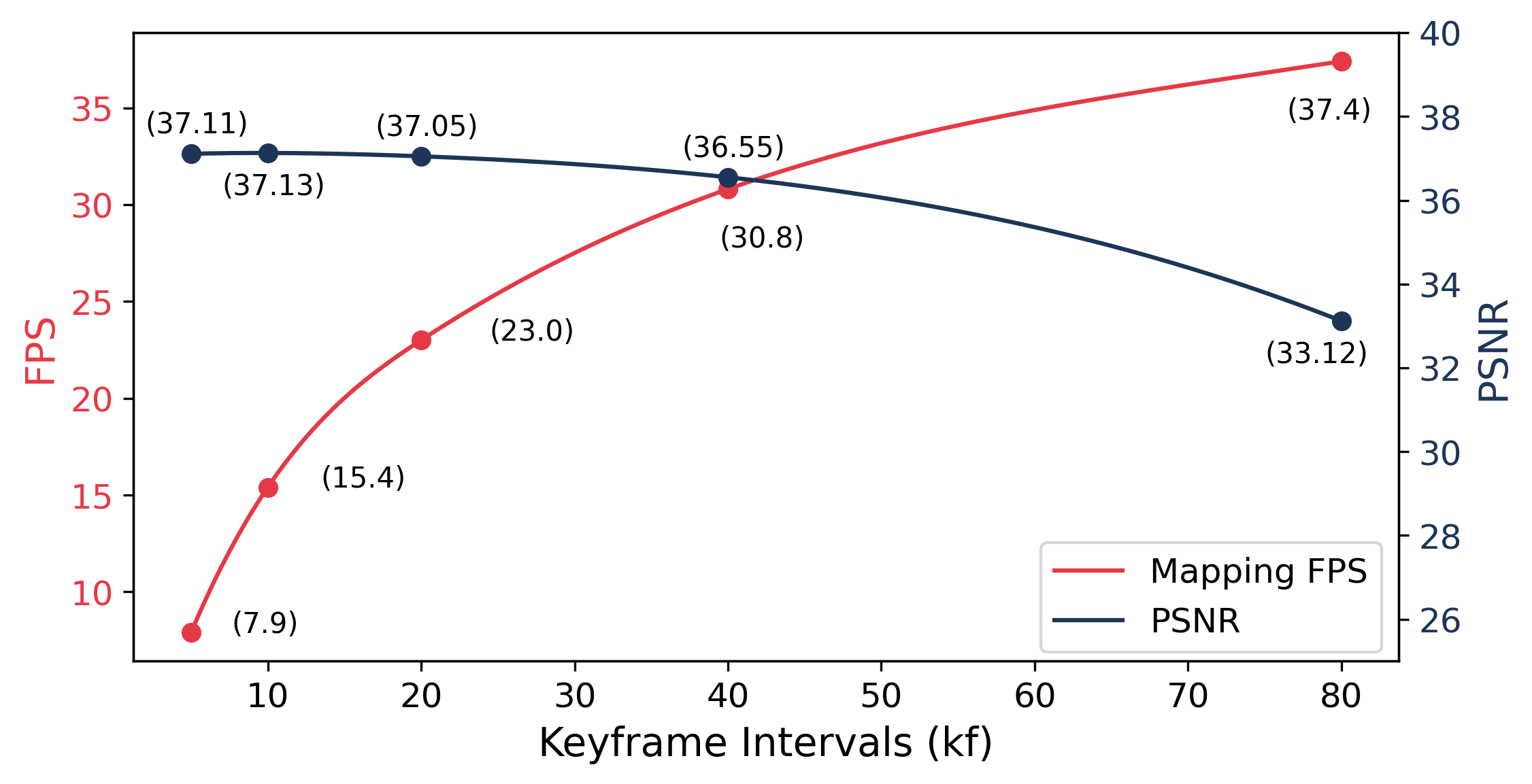}
\caption{Ablation study on keyframe intervals, presenting the mapping FPS and rendering PSNR as functions of keyframe intervals on the Room0 of the Replica dataset \cite{straub2019replica}.}
\label{fig:abaltion-keyframe}
\end{figure}

\noindent \textbf{Ablation of Keyframe Interval} As discussed in \Cref{sec:result-scene-densification}, the NeRF model provides exceptional interpolation capabilities and robust initialization, enabling the integration of a sparse keyframe list, which significantly reduces the computation time. However, excessively large intervals eventually result in insufficient supervision, thereby lowering reconstruction quality. \Cref{fig:abaltion-keyframe} examines the trend of mapping FPS and rendering PSNR in relation to keyframe intervals, showing a decline beyond an interval of 40. Consequently, we balance this trade-off by using a keyframe interval of 20 in our system. \label{sec:ablation-key-frame}

\begin{figure}[!tp]
\centering
\includegraphics[width=0.48\textwidth]{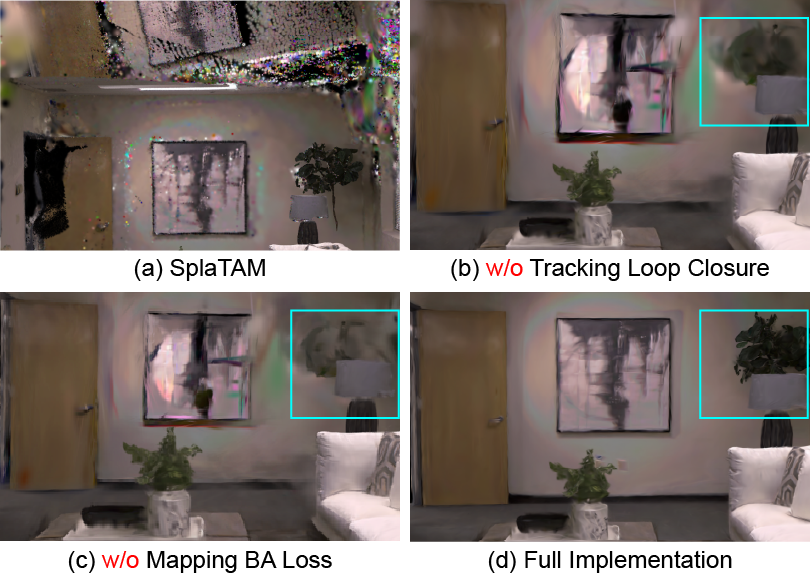}
\caption{Ablation study on BA, showing the rendered scenes of Apartment1 from the Replica dataset \cite{straub2019replica}. (a) baseline method employing naive frame-to-frame tracking. (b) no loop closure and BA on the tracking system. (c) no BA loss defined in \Cref{eq:ba-rgb} and \Cref{eq:ba-depth}. (d) Full implementation of DenseSplat.}
\label{fig:abaltion-ba}
\end{figure}

\noindent \textbf{Ablation of Bundle Adjustment} \Cref{fig:abaltion-ba} visualizes the rendering results with and without BA and the refinement using BA loss. In contrast to the baseline method that suffers from substantial scene drift, incorporating loop closure in tracking and BA-driven refinement in mapping successfully tackles this challenge. \\

\begin{figure}[!tp]
\centering
\includegraphics[width=0.48\textwidth]{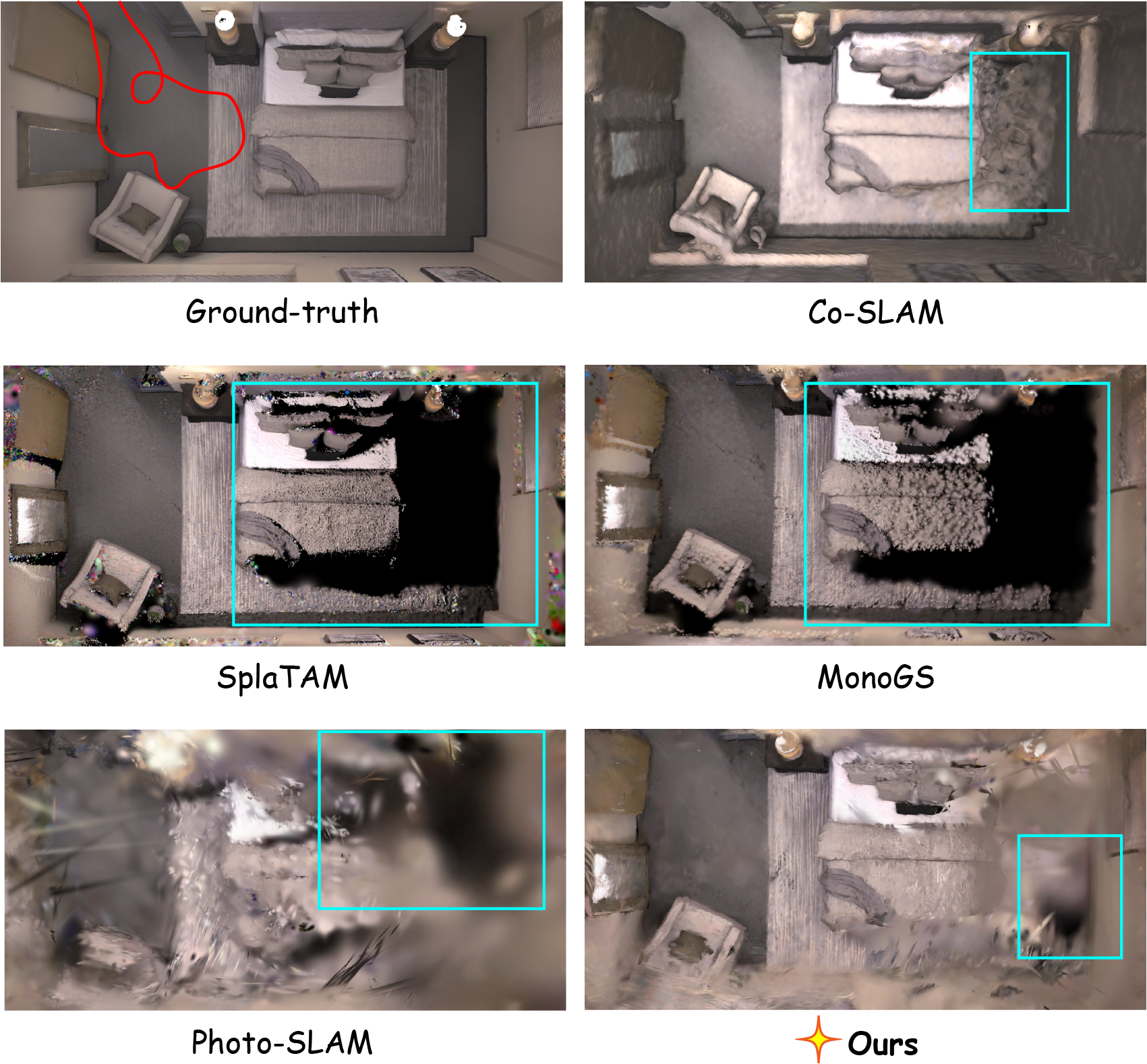}
\caption{Visualization of hole-filling failures in novel-view synthesis for the scene Apartment\_2 from the Replica dataset \cite{straub2019replica}. The camera trajectory is highlighted in red on the ground-truth map. Our method and Gaussian-based approaches struggle to completely fill the notably large gaps. In contrast, NeRF-based systems like Co-SLAM \cite{wang2023co} manage to fill these gaps to some extent, though the results are not sufficiently smooth.}
\label{sup-fig:failure-case}
\end{figure}

\noindent \textbf{Analysis of an Interpolation Failure Case} When the uncovered regions from viewpoints become extensively large, our method still faces challenges in completely filling holes. As illustrated in \Cref{sup-fig:failure-case}, in the case of Apartment\_2 from the Replica dataset \cite{straub2019replica}, the limitations become apparent when the camera trajectory does not sufficiently cover the scene, leaving nearly half of the room uncaptured. This situation makes scene interpolation particularly challenging. The NeRF-based model, such as Co-SLAM \cite{wang2023co}, manages to fill some gaps but introduces erroneous surfaces with significant artifacts. Our method outperforms other Gaussian-based methods by a large margin, using sparse primitives to fill gaps; however, it still leaves certain areas underrepresented. In scenarios where unobserved geometry primarily results from insufficient camera coverage, inferring these unobserved regions aligns more with scene extrapolation. Our method faces certain limitations as the NeRF prior is less reliable for extrapolation, and addressing these challenges will be a key direction for future research.

\section{VR/AR Applications}
\begin{figure}[!tp]
\centering
\includegraphics[width=0.42\textwidth]{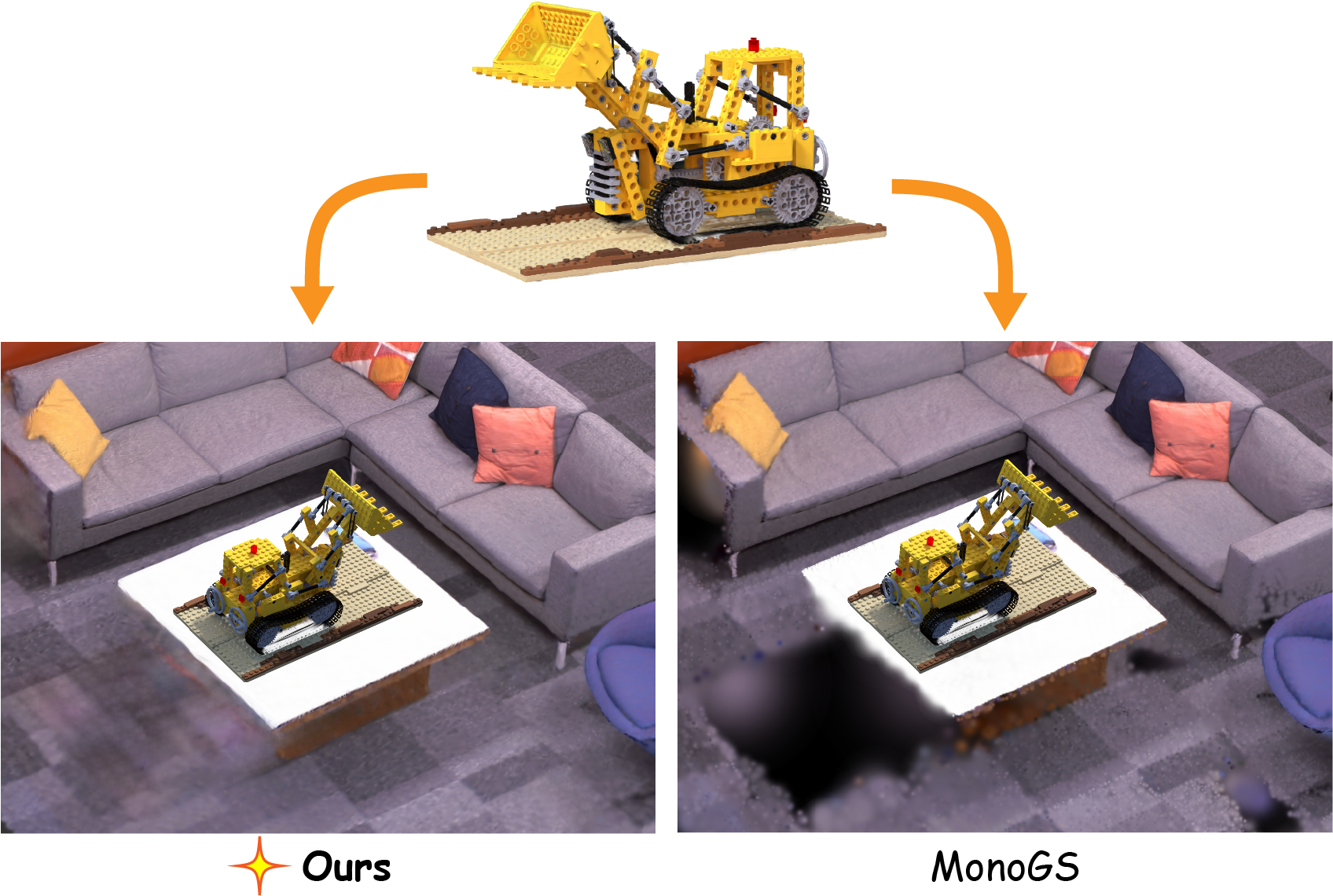}
\caption{Visulization of the VR/AR demo on Office\_3 of the Replica dataset \cite{straub2019replica}. We introduce the reconstructed LEGO object from the NeRF Synthetic Dataset \cite{mildenhall2021nerf} into the real-time reconstructed map and show the novel-view synthesis result.}
\label{fig:vr-ar}
\end{figure}

Neural dense SLAM is well-suited for VR/AR applications \cite{zhai2024nis}, where the essential synthesis of novel views enables seamless and dynamic perspectives that adapt to user movements. Despite the substantial advantages offered by Gaussian-based systems, such as producing high-fidelity maps and efficient real-time rendering capabilities, they often display significant holes and gaps when confronted with limited viewpoint supervision, which can severely impact the user experience. As illustrated in \Cref{fig:vr-ar}, we place the LEGO object reconstructed from the NeRF Synthetic Dataset \cite{mildenhall2021nerf} into the dense Gaussian map generated by SLAM systems. In comparison to previous methods like MonoGS \cite{matsuki2024gaussian}, our method remarkably improves scene completeness by interpolating missing geometry, thereby enhancing the user's immersion and interaction within the virtual environment.

\section{Conclusion}
We propose DenseSplat, the first visual dense SLAM system that seamlessly integrates the strengths of NeRF and 3DGS for robust tracking and mapping. DenseSplat targets practical challenges including obstructed views and the impracticality of maintaining dense keyframes, driven by hardware limitations and computational costs. This strategic integration enhances the system’s ability to interpolate missing geometries and robustly optimize Gaussian primitives with fewer keyframes, leading to fine-grained scene reconstructions and extraordinary real-time performance. Future research of this study could focus on implementing the system in practical mobile applications or multi-agent collaboration systems and conducting further experiments in real-world environments. \\

\noindent \textbf{Limitation} DenseSplat also has its limitations. As discussed in \Cref{sec:results_densification}, its scene interpolation capabilities are dependent on the NeRF model, and thus inherit NeRF's limitations. When the missing areas become excessively large and the NeRF model cannot adequately capture the geometry, both our method and NeRF struggle to extrapolate the scene geometry, leaving some regions underrepresented on the reconstructed map. Although recent efforts using generative priors show potential, extrapolation under realistic and large-scale conditions remains an open challenge. Additionally, because DenseSplat employs explicit Gaussian primitives for scene representation, storing the high-fidelity map requires more memory compared to NeRF models, which use neural implicit representations. Moreover, while DenseSplat supports submap systems to minimize memory usage during computation, there is room for a more advanced submap strategy that could potentially be expanded in real-world multi-agent systems. Addressing these limitations will be the focus of future research.

\bibliographystyle{IEEEtran}
\bibliography{ieee}

\begin{IEEEbiography}[{\includegraphics[width=1in,height=1.25in,clip,keepaspectratio]{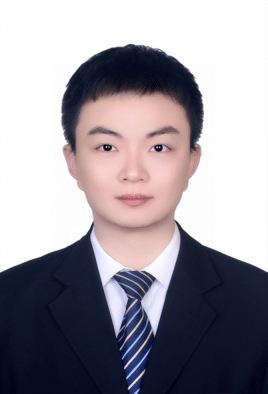}}]
{Mingrui Li} is currently pursuing the Ph.D. degree with the School of Information and Communication Engineering, Dalian University of Technology, Dalian, China. His research interests include 3D reconstruction, simultaneous localization and mapping (SLAM), and computer vision. \end{IEEEbiography}

\begin{IEEEbiography}[{\includegraphics[width=1in,height=1.25in,clip,keepaspectratio]{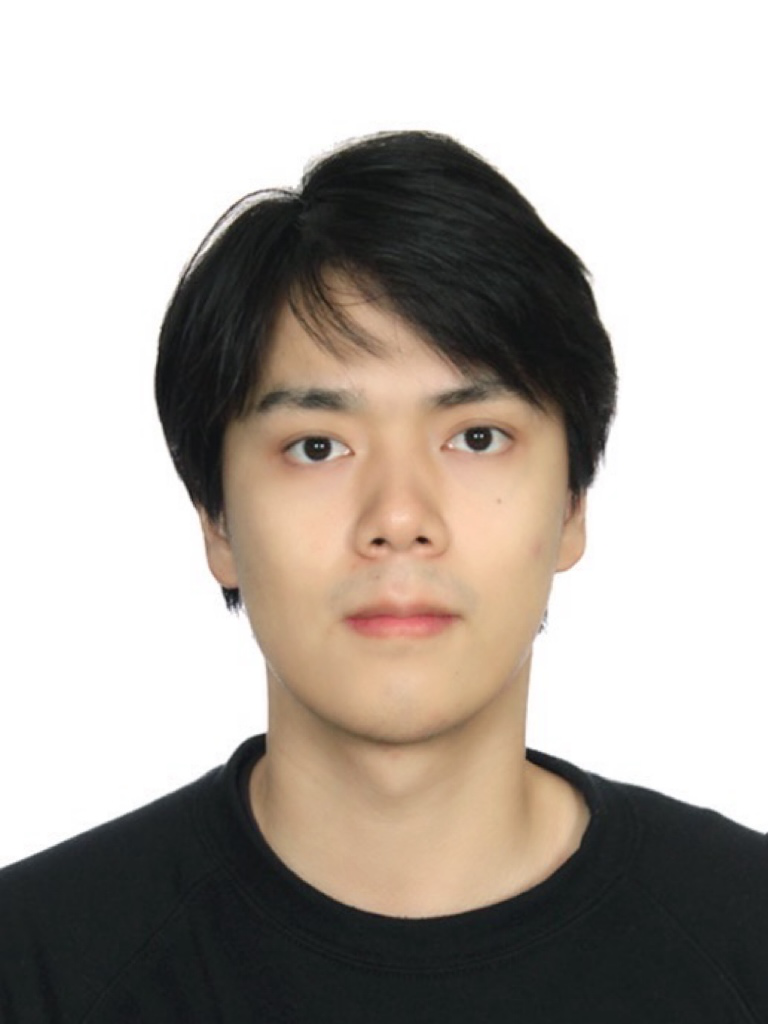}}]
{Shuhong Liu} is currently pursuing the Ph.D. degree in Department of Mechano-informatics, Information Science and Technology with the University of Tokyo, Tokyo, Japan. Before that, he received his bachelor's degree in Department of Electrical and Computer Engineering at University of Waterloo, Ontario, Canada, and the master's degree in Creative Informatics, Information Science and Technology, the University of Tokyo, Tokyo, Japan. His research interests include 3D computer vision, visual SLAM, and computational photography.
\end{IEEEbiography}

\begin{IEEEbiography}[{\includegraphics[width=1in,height=1.25in,clip,keepaspectratio]{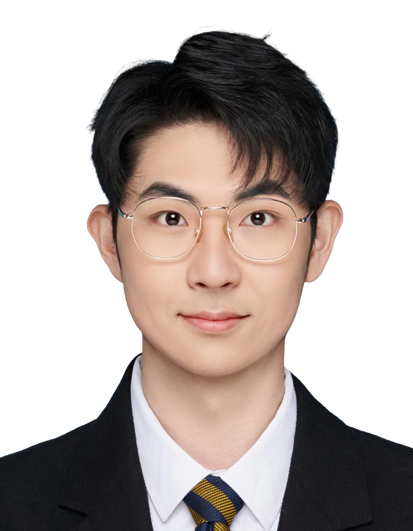}}]
{Tianchen Deng} is currently pursuing the
Ph.D. degree in control science and engineering with
Shanghai Jiao Tong University, Shanghai, China.
His main research interests include 3D Reconstruction, long-term
visual simultaneous localization and mapping
(SLAM), and vision-based localization. \end{IEEEbiography}

\begin{IEEEbiography}[{\includegraphics[width=1in,height=1.25in,clip,keepaspectratio]{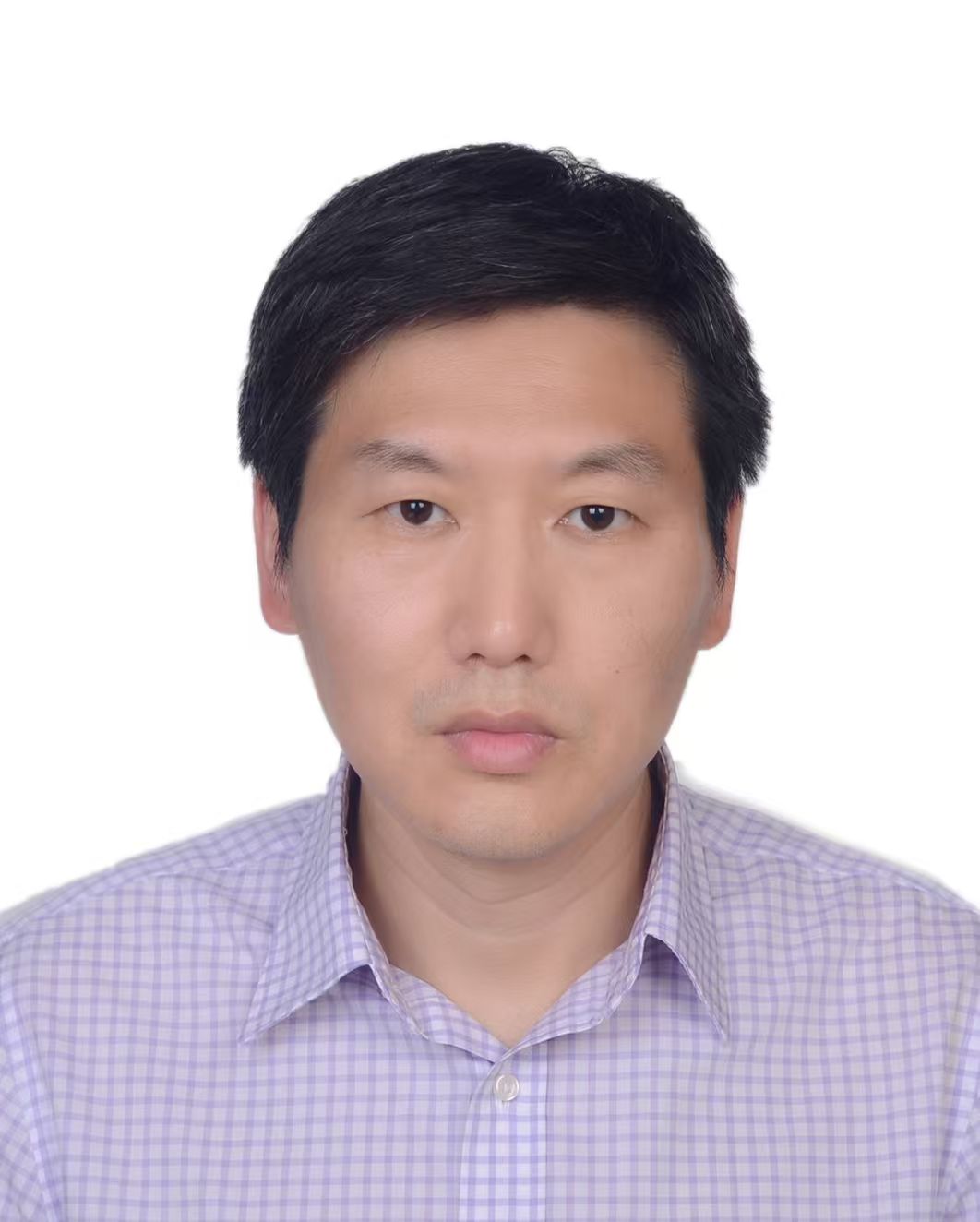}}]
{Hongyu Wang} (Member, IEEE) received his B.S. degree in electronic engineering from the Jilin University of Technology, Changchun, China, in 1990, the M.S. degree in electronic engineering from the Graduate School, Chinese Academy of Sciences, Beijing, China, in 1993, and the Ph.D. degree in precision instrument and optoelectronics engineering from Tianjin University, Tianjin, China, in 1997. He is currently a Professor with the Dalian University of Technology, Dalian, China. His research interests include image processing, computer vision, 3-D reconstruction, and simultaneous localization and mapping (SLAM). \end{IEEEbiography}

\end{document}